%% file: main.tex
\documentclass{article}
\usepackage{iclr2027_conference,times}

\input{math_commands.tex}

\usepackage[utf8]{inputenc} 
\usepackage[T1]{fontenc}    

\definecolor{cvprblue}{rgb}{0.21,0.49,0.74}
\usepackage[pagebackref=false,breaklinks=true,colorlinks,bookmarks=false,citecolor=cvprblue, linkcolor=cvprblue]{hyperref}

\usepackage{url}            
\usepackage{wrapfig}
\usepackage{booktabs}       
\usepackage{amsfonts}       
\usepackage{nicefrac}       
\usepackage{microtype}      
\usepackage{enumitem}       
\setlist[enumerate]{leftmargin=*} 
\usepackage{soul}
\usepackage{xcolor}         
\usepackage{subcaption}
\usepackage{graphicx}

\usepackage[most]{tcolorbox} 

\definecolor{jclightblue}{HTML}{C0CCFF}

\newcommand{\hlc}[2][yellow]{{\sethlcolor{#1}\hl{#2}}}

\newcommand{\tablestyle}[2]{\setlength{\tabcolsep}{#1}\renewcommand{\arraystretch}{#2}\centering\footnotesize}

\usepackage{multirow}
\usepackage[table]{xcolor}

\usepackage{tikz}
\usetikzlibrary{positioning, calc, fit, backgrounds, arrows.meta}

\definecolor{attblue}{HTML}{C6DBF7}   
\definecolor{moepeach}{HTML}{FDE0BC}  
\definecolor{embgray}{HTML}{EFEFEF}

\iclrfinalcopy
\title{
Let the Pixels Through: Rethinking Long Skip Connections for Scalable Pixel Diffusion
}

\author{%
  Jiachen Lei\qquad Jiahong Wu\thanks{Corresponding author} \qquad Xiangxiang Chu \\
  AMAP, Alibaba Group
}

\begin{document}
\maketitle

\begin{abstract}
Training pixel diffusion models remains less straightforward, owing to the ad hoc training recipes and architecture designs, along with the sensitivity to hyper-parameter choices.
We show that strong denoisers can be built by refining existing pixel-native designs from the literature.
Specifically, we revisit the long skip connection in U-Net and find that the optimization benefit commonly attributed to it is not necessary for its denoising benefit.
We further find that the conventional linear combination of features from the skip and main backbone is inadequate for regimes of various noise levels.
Motivated by these observations, we introduce two core modifications.
First, we adopt asymmetric long skips, which propagate fine-grained information from input embeddings to decoder layers.
Second, we introduce adaGLU, a SwiGLU-like MLP block, in decoder layers to adaptively filter information from the embeddings before being added to main-branch features.
The above designs give a simple pixel diffusion transformer architecture. 
When trained from scratch, it scales smoothly from small-scale baselines up to 1.6B parameters, achieving 1.62 and 1.70 FID on ImageNet-256 and -512, respectively.
Besides, it natively supports multimodal inputs, such as text and images, yielding competitive performance in text-to-image generation. We release our codes at \url{https://github.com/AMAP-ML/PiT}.
\end{abstract}

\begin{figure}[h]
    \centering
    \begin{minipage}[t]{.49\linewidth}
        \centering
        \includegraphics[width=\linewidth]{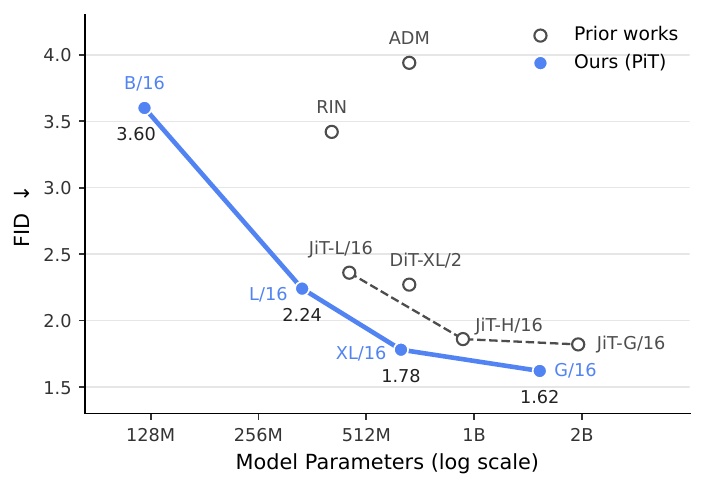}
    \end{minipage}
    \begin{minipage}[t]{.49\linewidth}
        \centering
        \includegraphics[width=\linewidth]{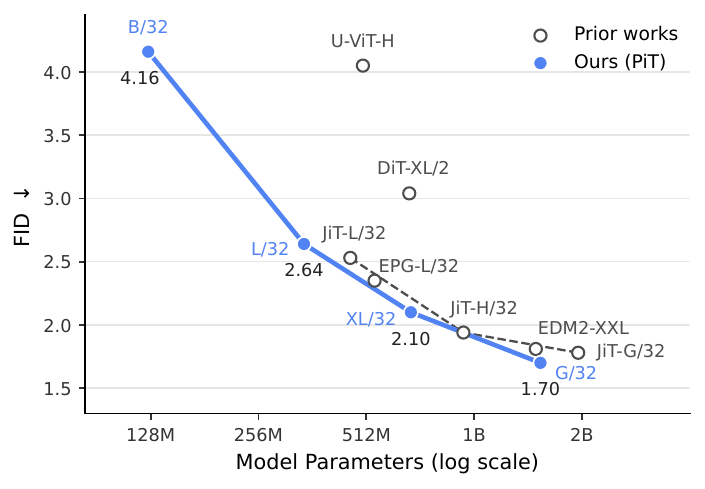}
    \end{minipage}
    \caption{Performance of our models on \textbf{(left)} ImageNet-256 and \textbf{(right)} ImageNet-512.
     We use identical training hyper-parameters across all model sizes and resolutions.}
    \label{fig:bubble}
\end{figure}

\input{model}
\input{introduction}
\input{related_work}
\input{background}
\input{method}
\input{experiment}
\section{Discussions \& Limitations}
\textbf{Discussions.}
U-Net was once a standard architecture for diffusion model.  Our study revisits its design from a different perspective. We find that the useful role of the long skip does not require its symmetric encoder-decoder structure. The skip can instead be viewed primarily as a pathway for low-level information, and its fusion with the backbone can be made adaptive to the denoising state.
This leads to a simpler formulation  in which the main backbone remains a standard transformer and the long skip is implemented as an asymmetric, noise-adaptive pathway. 

\textbf{Limitations.}
Our experiments are primarily conducted on ImageNet image generation. Although PiT scales to 1.6B parameters in our experiments, its behavior at substantially larger scales remains unclear.
We also did not evaluate PiT on video generation or other high-dimensional sequential data, where the benefit of the simplified routing design may slightly differ. Finally, while the token-based formulation suggests that PiT can be incorporated into systems beyond image generation, we have not evaluated adaGLU as a standalone component in existing large language model architectures.

\section{Acknowledgments}
We would like to sincerely thank Julius Berner for valuable suggestions throughout the course of this research, which helped us improve and strengthen this work.




\newpage
\appendix
\input{appendix}

\newpage
\bibliography{iclr2027_conference}
\bibliographystyle{iclr2027_conference}

\end{document}

%% file: math_commands.tex
\usepackage{amsmath,amsfonts,bm}

\def\eqref#1{equation~\ref{#1}}

\def\1{\bm{1}}

\def\vh{{\bm{h}}}

\def\vw{{\bm{w}}}
\def\vx{{\bm{x}}}

\def\vz{{\bm{z}}}

\DeclareMathAlphabet{\mathsfit}{\encodingdefault}{\sfdefault}{m}{sl}
\SetMathAlphabet{\mathsfit}{bold}{\encodingdefault}{\sfdefault}{bx}{n}



%% file: model.tex
\begin{figure}[h]
    \hspace{18pt}
    \begin{minipage}[b]{.58\linewidth}
    \centering
    \definecolor{gray_bbox_color}{RGB}{243,243,244}
    \definecolor{attn_color}{RGB}{155, 179, 255}
    \definecolor{moe_color}{RGB}{251, 209, 154}
    \definecolor{mix_path_color}{RGB}{130, 75, 80}
    \definecolor{attnres_purple}{RGB}{100,60,160}
    \definecolor{seagreen}{RGB}{235, 165, 85}
    \definecolor{attnres_box}{RGB}{212,228,200}
    \definecolor{adaln_color}{RGB}{235,210,245}
    \begin{subfigure}[b]{0.4\linewidth}
      \centering
      \resizebox{\linewidth}{!}{%
        \begin{tikzpicture}[
          emb/.style={
              rectangle,
              rounded corners=2pt,
              minimum width=52pt,
              minimum height=8pt,
              draw=black,
              semithick,
              fill=gray_bbox_color,
              font=\scriptsize
            },
          blockbox/.style={
              rectangle,
              rounded corners=3pt,
              minimum width=58pt,
              minimum height=11pt,
              draw=black,
              semithick,
              fill=attn_color,
              font=\scriptsize
            },
          dotbox/.style={
              rectangle,
              minimum width=58pt,
              minimum height=8pt,
              font=\footnotesize
            },
          outerblock/.style={ rectangle, rounded corners=6pt, draw=black, dashed, semithick, inner sep=5pt },
          residual/.style={
              -{Latex[length=3.5pt, width=2.5pt]},
              line width=0.6pt,
              rounded corners=3pt,
              black,
              shorten <=0.5pt,
              shorten >=0.5pt,
            },
          mainlink/.style={
              -{Latex[length=3.5pt, width=2.5pt]},
              line width=0.5pt,
              rounded corners=3pt,
              black,
              shorten <=0.2pt,
            },
          ]
    
          \path (-30pt,0) -- (30pt,0);
    
          \node[emb] (emb) at (0,0) {Embedding};
    
          \node[blockbox, anchor=south, yshift=14pt]
            at (emb.north) (block1) {Block};
    
          \node[blockbox, anchor=south, yshift=10pt]
            at (block1.north) (block2) {Block};
    
          \node[blockbox, anchor=south, yshift=10pt]
            at (block2.north)  (block3) {Block};
    
          \node[dotbox, anchor=south, yshift=10pt]
            at (block3.north)  (dots1) {$\vdots$};
    
    
        \node[blockbox, anchor=south, yshift=10pt]
            at (dots1.north)  (block6) {Block};
    
        \node[blockbox, anchor=south, yshift=10pt]
            at (block6.north) (block7) {Block};
    
        \node[blockbox, anchor=south, yshift=10pt]
            at (block7.north) (block8) {Block};
    
          \node[emb, anchor=south, yshift=28pt,
            draw=none, fill=none]
            at (block8.north) (output) {Output};
    
          \draw[mainlink] (emb.north) -- (block1.south);
          \draw[mainlink] (block1.north) -- (block2.south);
          \draw[mainlink] (block2.north) -- (block3.south);
          \draw[mainlink] (block3.north) -- (dots1.south);
          \draw[mainlink] (dots1.north) -- (block6.south);
          \draw[mainlink] (block6.north) -- (block7.south);
          \draw[mainlink] (block7.north) -- (block8.south);
          \draw[mainlink] (block8.north) -- (output.south);
    
        \coordinate (bbot) at ([yshift=-6pt]emb.south);
        \coordinate (btop) at ($(block8.north)!0.65!(output.south)$);
        \coordinate (bleft) at ([xshift=-38pt]block1.center);
        \coordinate (bright) at ([xshift=38pt]block1.center);
        \begin{scope}[on background layer]
            \node[outerblock, inner sep=0pt, fit= (bbot -| bleft)(btop -| bright)] (mainblock) {};
        \end{scope}
          \coordinate (lx) at ([xshift=26pt]block1.east);
    
          \draw[residual]
            (block1.east)
            -- (lx |- block1.center)
            -- (lx |- block8.center)
            -- (block8.east);
    
          \coordinate (lx2) at ([xshift=23pt]block2.east);
    
          \draw[residual]
            (block2.east)
            -- (lx2 |- block2.center)
            -- (lx2 |- block7.center)
            -- (block7.east);
    
          \coordinate (lx3) at ([xshift=20pt]block3.east);
    
             \draw[residual]
            (block3.east)
            -- (lx3 |- block3.center)
            -- (lx3 |- block6.center)
            -- (block6.east);
    
    
    
        \end{tikzpicture}%
      }
      \caption{\scriptsize U-Net-style long skips.}
      \label{fig:vit-unet-skips}
    \end{subfigure}
    \hfill
    \begin{subfigure}[b]{0.42\linewidth}
      \centering
      \resizebox{\linewidth}{!}{%
        \begin{tikzpicture}[
          emb/.style={
              rectangle,
              rounded corners=2pt,
              minimum width=52pt,
              minimum height=8pt,
              draw=black,
              semithick,
              fill=gray_bbox_color,
              font=\scriptsize
            },
          blockbox/.style={
              rectangle,
              rounded corners=3pt,
              minimum width=58pt,
              minimum height=11pt,
              draw=black,
              semithick,
              fill=attn_color,
              font=\scriptsize
            },
           adaglu/.style={
              rectangle,
              rounded corners=3pt,
              minimum width=58pt,
              minimum height=11pt,
              draw=black,
              semithick,
              fill=moe_color,
              font=\scriptsize
            },
          dotbox/.style={
              rectangle,
              minimum width=58pt,
              minimum height=8pt,
              font=\footnotesize
            },
          outerblock/.style={ rectangle, rounded corners=6pt, draw=black, dashed, semithick, inner sep=5pt },
          residual/.style={
              -{Latex[length=3.5pt, width=2.5pt]},
              line width=0.6pt,
              rounded corners=3pt,
              black,
              shorten <=0.5pt,
              shorten >=0.5pt,
            },
          mainlink/.style={
              -{Latex[length=3.5pt, width=2.5pt]},
              line width=0.5pt,
              rounded corners=3pt,
              black,
              shorten <=0.2pt,
            },
            condlink/.style={
                -{Latex[length=3.5pt, width=2.5pt]},
                line width=0.5pt,
                rounded corners=3pt,
                black!75,
                shorten <=0.2pt, 
            },
          ]
    
          \path (-30pt,0) -- (30pt,0);
    
          \node[emb] (emb) at (0,0) {Embedding};
    
          \node[blockbox, anchor=south, yshift=14pt]
            at (emb.north) (block1) {Block};
    
          \node[blockbox, anchor=south, yshift=10pt]
            at (block1.north) (block2) {Block};
    
    
          \node[dotbox, anchor=south, yshift=10pt]
            at (block2.north)  (dots1) {$\vdots$};
    
    
        \node[adaglu, anchor=south,  yshift=10pt]
            at (dots1.north)  (adaglu1) {adaGLU};
    
        \node[blockbox, anchor=south, yshift=10pt]
            at (adaglu1.north)  (block6) {Block};
    
        \node[adaglu, anchor=south,  yshift=10pt]
            at (block6.north)  (adaglu2) {adaGLU};
    
        \node[blockbox, anchor=south, yshift=10pt]
            at (adaglu2.north) (block7) {Block};
    
    
    
        \node[emb, anchor=south, yshift=28pt,
            draw=none, fill=none]
            at (block7.north) (output) {Output};
    
        \node[emb, minimum width=18pt] (emb1)
          at ([xshift=50pt]adaglu1.center) {Emb.};
    
        \node[emb, minimum width=18pt] (emb2)
          at ([xshift=50pt]adaglu2.center) {Emb.};
        
    
        \draw[condlink] (emb1.west) -- (adaglu1.east);
        \draw[condlink] (emb2.west) -- (adaglu2.east);
    
          \draw[mainlink] (emb.north) -- (block1.south);
          \draw[mainlink] (block1.north) -- (block2.south);
          \draw[mainlink] (block2.north) -- (dots1.south);
          \draw[mainlink] (dots1.north) -- (adaglu1.south);
          \draw[mainlink] (adaglu1.north) -- (block6.south);
          \draw[mainlink] (block6.north) -- (adaglu2.south);
          \draw[mainlink] (adaglu2.north) -- (block7.south);
          \draw[mainlink] (block7.north) -- (output.south);
    
        \coordinate (bbot) at ([yshift=-6pt]emb.south);
        \coordinate (btop) at ($(block8.north)!0.65!(output.south)$);
        \coordinate (bleft) at ([xshift=-38pt]block1.center);
        \coordinate (bright) at ([xshift=38pt]block1.center);
        \begin{scope}[on background layer]
            \node[outerblock, inner sep=0pt, fit= (bbot -| bleft)(btop -| bright)] (mainblock) {};
        \end{scope}
    
        \end{tikzpicture}%
      }
      \caption{\scriptsize Ours.\ \ \ \ \ \ \ \ \ \ \ }
      \label{fig:pit}
    \end{subfigure}
    \end{minipage}
    \hfill
    \begin{minipage}[b]{.28\linewidth}
        \includegraphics[width=.85\linewidth]{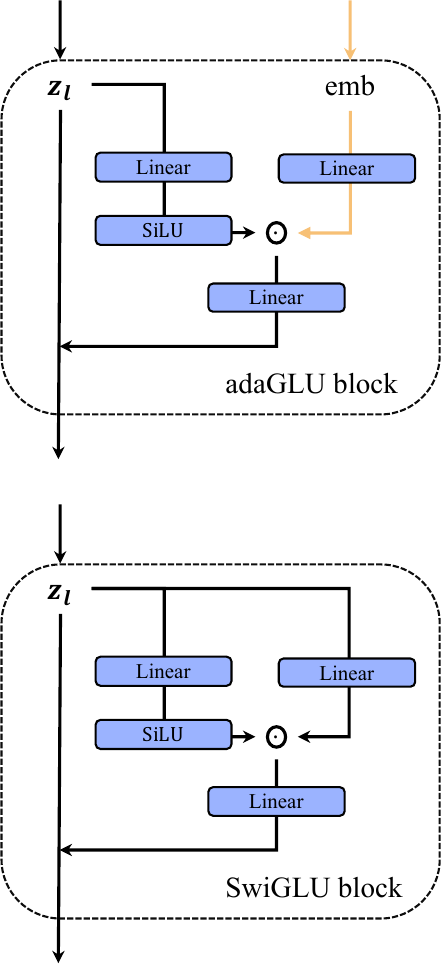}
        \subcaption[figure]{\scriptsize Comparing adaGLU with SwiGLU.}
        \label{fig:adaglu_vs_swiglu}
    \end{minipage}
    \caption{Overview of our model architecture. In \textbf{(a, b)}, we compare our architecture against the standard vision transformer (ViT) model with U-Net-style long-skip connections. The embeddings input to the adaGLU block are the same as that of the first ViT block. \textbf{(c)} we show that an adaGLU block shares the same architecture as a standard SwiGLU block.}
\end{figure}

%% file: introduction.tex
\section{Introduction}
Pixel-space diffusion models have recently received renewed attention.
A key advantage is that they avoid the computational overhead brought by a separate tokenizer in latent diffusion models (LDM)~\citep{rombach2022highresolutionimagesynthesislatent}, thus being attractive for high-dimensional generation.
Recent work has shown that strong pixel-space diffusion can be obtained with relatively simple transformer backbones~\citep{vaswani2017attention, dosovitskiy2021imageworth16x16words}.
Some methods~\citep{li2025basicsletdenoisinggenerative} train directly from pixels, but are sensitive to their architectural design and other hyper-parameters.
The other~\citep{yu2026pixelditpixeldiffusiontransformers, lei2026vaeendtoendpixelspacegenerative} further introduce additional training stages or specialized architecture components.
These designs achieve strong performance, but also raise a basic question: how much architectural structure is actually necessary for a strong pixel-space diffusion model?

We revisit a basic component, the long skip connection in U-Net~\citep{ronneberger2015unetconvolutionalnetworksbiomedical}, a common design in diffusion models. 
In its standard form, features from shallow encoder layers are directly connected to corresponding decoder layers. 
The motivation is straightforward. Shallow features preserve fine-grained information that may otherwise be lost after passing through a deep semantic backbone.
These long residuals are also commonly believed to facilitate optimization by providing a shortcut for gradient propagation.
In U-Net-style designs, these two effects are coupled by the symmetric encoder-decoder topology. Moreover, skip features are typically concatenated with the backbone features and fused by a fixed linear projection.
In this paper, we investigate how much of these structures are actually necessary for denoising, and suggest that a strong denoiser can be achieved with an improved design.

We first examine whether the optimization effect is essential.
In specific, we detach the gradient pass through the skip connections and find that it does not actually hurt performance. 
This reveals that, to retain the benefit of the skip, the conventional pairwise encoder-decoder structure is not always necessary in all situations. In practice, the optimization issue only becomes visible when we increase model depth. However, we show that this issue can be properly addressed with standard weight initialization.
Motivated by the gradient-detach experiment, we then remove the pairwise encoder-decoder connections and route the input embedding directly to decoder layers. Surprisingly, this asymmetric design further improves performance. 
These observations together suggest that optimization benefit commonly attributed to long skips is not essential for their denoising benefit.
The long skip connections are better understood as a pathway for low-level information than as a solution for optimization difficulty.

Then, this raises another question about how the skip features should be combined with the deep features in the backbone. In diffusion, the answer may depend on the noise level. At low noise, the input retains considerable image information and is useful for reconstruction. 
At high noise, the same information is less reliable.  
Concatenation followed by a fixed linear layer, as in common practice, does not distinguish these cases. It lacks a basic gating mechanism to adapt to regimes at various noise levels, yet directly merges features by channel-wise linear combination. 
We therefore test nonlinear gating functions and find that a SwiGLU-style formulation~\citep{shazeer2020gluvariantsimprovetransformer, dauphin2017languagemodelinggatedconvolutional} gives a substantial improvement. Motivated by this result, we repurpose the standard SwiGLU as a gate module, which we call \textit{adaGLU}, for features from the long skip pathway. Figure~\ref{fig:adaglu_vs_swiglu} compares its structure against a standard SwiGLU block. Specifically, adaGLU rectifies the skip feature with the activation from the main backbone. The skip feature provides low-level information, while the main-branch feature controls how much of this information is used. 
After this, we fuse features by simple addition.

The above modifications give a simple transformer architecture, which we call Pixel Diffusion Transformer \textbf{(PiT)}. We provides an overview of its structure in Figure~\ref{fig:pit}. 
We mainly focus on identifying necessary components for a scalable architecture prototype. We follow this philosophy throughout our work, and refrain from introducing additional network designs or training techniques.
As a result, our model \textit{relies on no tokenizer, no AdaLN, no in-context embedding, no pre-training, and no auxiliary loss}.

We conduct experiments on ImageNet-256 and ImageNet-512, referring to ImageNet-1K~\citep{deng2009imagenet} at $256\times256$ and $512\times512$ resolution, respectively.
When trained from scratch, it reaches 1.62 FID on ImageNet-256 and 1.70 FID on ImageNet-512.
Besides, our network also natively supports multi-modal inputs by treating them uniformly as a token sequence.
To show this, we train a text-to-image generation model at 512$\times$512 resolution by concatenating text embeddings with image tokens as model input.
We evaluate it on GenEval~\citep{ghosh2023genevalobjectfocusedframeworkevaluating} and DPG-Bench~\citep{hu2024ellaequipdiffusionmodels}, scoring 0.82 and 84.3, respectively. 

Our observations do not imply that U-Net-style architectures are ineffective.
Rather, they suggest that their structural assumptions are not necessary for retaining the useful function of long skips. 
We summarize our contributions below.
\begin{enumerate}
\item  We revisit the role of long skip connections in pixel diffusion and show that their denoising benefit can be separated from their advantage in optimization, allowing the conventional U-Net topology to be replaced by a simpler asymmetric information pathway. We reinterpret the long skip connection as a functional component rather than an architectural template.
\item We strengthen its feature fusion mechanism by introducing adaGLU. It repurposes the standard SwiGLU as a gating module, allowing the backbone features to modulate the low-level information carried by the skip.This design benefits the model with long-skip connections in both pixel- and latent-space training.
\item We show that PiT performs strongly when trained from scratch across a wide range of model sizes, reaching 1.62 FID on ImageNet-256 and 1.70 FID on ImageNet-512. We further show that the same architecture naturally extends to text-to-image generation.
\end{enumerate}

%% file: related_work.tex
\section{Related Work}

\textbf{Pixel-space Diffusion Model} have recently regained attention. By removing the pretrained VAE tokenizer, they avoid the reconstruction bottleneck, while offering better efficiency in direct image modeling.
Recent works can be roughly divided into two categories. The first group uses specialized training recipes~\citep{chen2025pixelflowpixelspacegenerativemodels, wang2025pixnerdpixelneuralfield, hoogeboom2023simplediffusionendtoenddiffusion, lei2023masked}. EPG~\citep{lei2026vaeendtoendpixelspacegenerative} and USP~
\citep{chu2025usp} introduces self-supervised pretraining before diffusion training.
Though achieving impressive performance, they suggest that competitive pixel diffusion often depends on extra training objectives/stages.
In contrast, JiT~\citep{li2025basicsletdenoisinggenerative} offers a workaround by reshaping the DiT~\citep{peebles2023scalablediffusionmodelstransformers} architecture and parameterizing the denoising model to directly predict clean images. When trained from scratch, it yields competitive performance to top-performing models. However, it is sensitive to hyper-parameter choices, such as in-context embedding and Dropout~\citep{JMLR:v15:srivastava14a}. In comparison, our model requires no specialized recipe, presenting clean scaling behavior without extensive hyper-parameter tuning.

\textbf{Denoising Network.} The diffusion model architecture has undergone a series of advancements from U-Net like CNNs~\citep{song2021scorebasedgenerativemodelingstochastic, ho2020denoisingdiffusionprobabilisticmodels, dhariwal2021diffusionmodelsbeatgans} to transformer-based architecture~\citep{peebles2023scalablediffusionmodelstransformers, bao2023worthwordsvitbackbone, jabri2023scalableadaptivecomputationiterative, tian2024uditsdownsampletokensushaped}.
Recently, a notable trend is to incorporate shallow-to-deep information routing, such as  DDT~\citep{wang2025ddtdecoupleddiffusiontransformer}, and PixelDiT. Such designs can be effective, but they also introduce additional architectural structure beyond a standard transformer.
Another work, DAR~\citep{xu2026rethinkingcrosslayerinformationrouting}, employs attention residual~\citep{kimiteam2026attentionresiduals} to replace the fixed U-Net-style skip connections with learnable long residuals. The drawback is the additional heavy cost of computing the adaptive weights for each residual outweighs the benefits, showing a poor trade-off. 
Our work examines how much architectural structure that pathway actually needs and how should skip features be fused with backbone features, given that shallow-to-deep information routing is useful.
Empirically, our designs could also benefit model with long-skip connections in both pixel- and latent-space training (see Table~\ref{tab:ablate_skip} and~\ref{app_tab:latent_space}).

%% file: background.tex
\section{Background}

Diffusion Models~\citep{ho2020denoisingdiffusionprobabilisticmodels, song2021scorebasedgenerativemodelingstochastic} are a class of generative models that learn to generate data by reversing a pre-defined forward diffusion process. Given a noise distribution of standard Gaussian distribution $\bm{\epsilon}\sim\mathcal{N}(0,\bm{I})$ and the target data distribution $p_{data}(\vx)$, the forward process traverses the data towards the noise distribution  
We adopt the diffusion model of variance-exploding (VE) forward SDE defined in EDM~\citep{karras2022elucidating}: $d\vx=\sqrt{2t}d\bf{w}$. Thus, the transition kernel from clean image $\vx(0) \sim p_{data}(\vx)$ to $\vx(t)$ is defined as $p(\vx(t)|\vx(0))=\mathcal{N}(\vx(t); \vx(0), t^2\bm{I})$, $t\in[\sigma_{min}, \sigma_{max}]$. We choose $\sigma_{min}=0.002$ and $\sigma_{max}$ = $80$. 
When $t=\sigma_{max}$, $\vx(t)$ approximates the noise distribution $\mathcal{N}(0, \sigma_{max}^2\bm{I})$. The reverse process of the variance-exploding forward SDE is defined as
\begin{equation}
    \label{eq:reverse}
    d\vx = [- 2t\nabla_{\vx}\log p(\vx(t))]dt + \sqrt{2t}d\mathbf{\bar{w}}.
\end{equation}
Here, $\nabla_{\vx}\log p(\vx(t))$ is the score function, and $\bar{\vw}$ is the standard Wiener process that flows backward in time.
The diffusion model trains a denoiser $f_\theta$ to approximate a reparameterization of $\nabla_{\vx}\log p(\vx(t))$ in the pre-defined time interval. In specific, it's optimized with a simple MSE loss
\begin{equation}
    \mathcal{L}=\mathbb{E}_{t,\vx(0), \vx(t)}\biggl[w(t)\|\vx(0)-f_{\theta}(\vx(t),t)\|_2^2\biggr].
    \label{eq:denoising_objective}
\end{equation}  
$w(t)$ is a scalar weighting function.
We adopt the above $\vx$-prediction formulation throughout our work.
To generate images, starting from $t=\sigma_{max}$, we iteratively solve the corresponding probability flow ordinary differential equation (PF ODE)~\citep{song2021scorebasedgenerativemodelingstochastic} of the reverse diffusion process in \eqref{eq:reverse} with ODE solvers such as Euler-Maruyama or Heun~\citep{karras2022elucidating}.
The PF ODE is depicted by
\begin{equation}
     d\vx = [-t\nabla_{\vx}\log p(\vx(t))]dt.
\end{equation}

%% file: method.tex
\section{Method}
\begin{table}
    \centering
    \caption{We compare representative skip connection designs of prior diffusion models, including EDM2~\citep{Karras2024edm2}, Simple Diffusion (SD)~\citep{hoogeboom2023simplediffusionendtoenddiffusion}, and UniDiffuser~\citep{bao2023one}. We omit bias terms for simplicity. 
    $\vh, \vz\in\mathbb{R}^d$ are features from the skip connection and the backbone, respectively.
    $\bm{W} \in \mathbb{R}^{d\times 2d}$ is a linear layer. 
    LN is layer normalization~\citep{ba2016layernormalization}. }
    \scriptsize
    \begin{tabular}{c c c c c}
        \toprule
        \tablestyle{3pt}{1.1}
        &  EDM2 & SD & UniDiffuser & \textbf{Ours} \\
        \midrule
        Gating $\bm{G}(\vh, \vz)$ & $\vh$ &  $\vh$ & $\vh$ & $\text{adaGLU}(\vh, \text{RMSNorm}(\vz))$ \\
        Fusion $\bm{F}(\vh, \vz)$ & $\bm{W}\text{Concat}(\bm{h}, \vz)$ & $\frac{1}{\sqrt{2}}(\text{LN}(\bm{h}) + \text{LN}(\vz))$ & $\text{LN}(\bm{W}\text{Concat}(\bm{h}, \vz))$ & $\bm{G}(\vh, \vz) + \vz$ \\
        Skip topology & Symmetric & Symmetric & Symmetric & Asymmetric \\
        Backbone & CNN & Hybrid & ViT & ViT \\
        \bottomrule
    \end{tabular}
    \label{tab:compare_skip_connection}
\end{table}

Starting from a simple experiment that detaches the gradient from the skip connections, we examine how the skip connection design can be improved, and subsequently investigate the optimization issues that arise when scaling up the model.
\subsection{Problem Formulation}
Let $\vh, \vz \in \mathbb{R}^{d}$ denote the feature from the skip connection and the feature of the main branch respectively. The feature fusion mechanism in a network equipped with U-Net-style skip connections can be generally depicted as
\begin{equation}
    \vz = \bm{F}(\bm{G}(\vh, \vz), \vz).
\end{equation}
Here, $\bm{G}$ is the gating function and  $\bm{F}$ is the fusion operator that combines the gated output and $\vz$.
In Table~\ref{tab:compare_skip_connection}, we list the designs of skip connections in prior works. 
It suggests that they often use an identity gating function $\bm{G}(\vh, \vz)=\vh$ and a linear fusion operator that combines $\vh$ and $\vz$ via channel-wise linear combination (concatenation + linear map). They leave the deep merging of features to the main backbone, thus paying less attention to the design of the fusion mechanism itself.

\textbf{Settings.} In discussions below, we adopt the design in ~\citet{bao2023one} (third column in Table~\ref{tab:compare_skip_connection}) as a starting point.
By default, we use a 12-layer 768-dim ViT-B with a large patch size of 16$\times$16 as the base model. Images, class labels, and time conditioning are treated equally as input tokens.
We train all models for 200K steps with a batch size of 1024 on ImageNet-256. We report FID measured on 10k samples produced by an Euler ODE sampler with 50 sampling steps.

\subsection{Streamline the Skip Connection Design}
A common belief is that U-Net-style skip connections facilitate training by alleviating the gradient vanishing problem.
To test this hypothesis, we compare two ViT-B models, both equipped with U-Net-style skip connections. One is trained as usual, while the other is trained by detaching the gradient from the skip connections. This forces the supervisory signal to flow exclusively backward through the main backbone.  Surprisingly, the simple modification yields a slightly better FID score than the baseline (50.1 v.s. 52.1), suggesting that the commonly assumed role of skip connections in mitigating gradient vanishing may not be necessary in this scenario.

Based on this observation, we then remove the pair-wise encoder-decoder shortcuts and redistribute the skip connections so that only the input embeddings are input to decoder layers. 
This retains the shortcuts to preserve fine-grained details from the input.
As shown in Table~\ref{tab:ablate_skip}, this asymmetric design improves the FID from 52.1 to 47.8. 
These results indicate that the primary function of the long skip is to provide an information pathway from the input toward the deeper layers, rather than to facilitate optimization.
This perspective also motivates a reconsideration of how skip features should be fused with the backbone.

\begin{table}
    \begin{minipage}[t]{.43\linewidth}
        \centering
        \caption{Improve skip connection design.}
        \label{tab:ablate_skip}
        \begin{tabular}{l c}
             & FID \\
            \toprule
            JiT-B\footnotemark & 38.1 \\
            ViT-B & \\
            \quad + symmetric skips & 52.1 \\
            \qquad + detach grad. & 50.1 \\
            \qquad + adaGLU & 41.2 \\
            \quad + asymmetric skips & 47.8 \\
            \rowcolor{jclightblue} \qquad + adaGLU & \textbf{36.2} \\ 
        \end{tabular}
    \end{minipage}
    \hfill
    \begin{minipage}[t]{.55\linewidth}
        \centering
        \caption{Ablate gating function $G(\vh, \vz)$. $\bm{W}_{bottleneck}\in\mathbb{R}^{d'\times d}$ and $d'<d$. $\bm{W}_1$ maps features back to $\mathbb{R}^d$, $\bm{W}_2, \bm{W}_3 \in \mathbb{R}^{D\times d}$ and $D>d$.}
        \label{tab:ablate_gating}
        \begin{tabular}{l c}
            $G(\vh, \vz)$ & FID \\
            \toprule
            $\bm{W}_{bottleneck}\vh$ & 54.1 \\
            $\bm{W}_1(\vh\odot\text{SiLU}(\vz))$ & 57.1 \\
            $\bm{W}_1(\bm{W}_2\vh\odot\text{Tanh}(\bm{W}_3\vz))$ &  39.0 \\
            $\bm{W}_1(\bm{W}_2\vh\odot\text{Sigmoid}(\bm{W}_3\vz))$ & 37.6 \\
             \rowcolor{jclightblue} $\bm{W}_1(\bm{W}_2\vh\odot\text{SiLU}(\bm{W}_3\vz))$ & \textbf{36.2} \\
        \end{tabular}
    \end{minipage}
\end{table}
\footnotetext{We reproduce with official implementation (https://github.com/LTH14/JiT), while replacing the fp32 attention with flash-attn in bf16 for training efficiency.}

Denoising fundamentally operates across varying noise levels, and the usefulness of the input signal varies accordingly.
At low noise levels, the perturbed input $\vx_t$ still preserves substantial image information and can therefore directly support image reconstruction.
In this regime, routing this information directly into deep layers can reduce the burden on the backbone to reproduce an identity-like mapping.
However, at high noise levels, the input is heavily corrupted and provides substantially less reliable information. 
In this regime, the network should be able to suppress the contribution of the skip pathway when it is not useful. 

To implement such adaptive routing, we compare several gating functions $\bm{G}(\vh, \vz)$ in Table~\ref{tab:ablate_gating}, ranging from a bottleneck linear layer to nonlinear MLP-based designs.
The bottleneck linear layer reduces the dimensionality of the skip features as an information bottleneck, whereas the nonlinear variants allow the skip representation to be modulated by the semantics encoded in the backbone features.
Among them, the SwiGLU formulation achieves the lowest FID.  Specifically, we define
\begin{equation}
    \text{adaGLU}(\vh,\vz)=\bm{W}_{1}(\bm{W}_{2}\vh\odot\text{SiLU}(\bm{W}_{3}\vz)).
    \label{eq:adaglu}
\end{equation}
Here, $\bm{W}_1\in \mathbb{R}^{d\times D}$, $\bm{W}_2\in \mathbb{R}^{D\times d}$, $\bm{W}_3\in \mathbb{R}^{D\times d}$, $D>d$. Based on our current settings in Table~\ref{tab:ablate_skip}, the feature fusion with adaGLU can be depicted by
\begin{equation}
    \vz =\operatorname{Norm}\biggl(\bm{W}\text{Concat}\bigl(\text{adaGLU}(\vh, \vz), \vz\bigr)\biggr).
    \label{eq:postnorm_skip}
\end{equation}
$\bm{W}\in \mathbb{R}^{d\times 2d}$. We name this formulation as Post-Norm adaGLU as it mirrors the structure of a Post-Norm ViT block, where the normalization layer is applied at the end of computation. 
The term $\text{SiLU}(W_3\vz)$ provides a semantic-dependent gate that modulates the information carried by the skip feature $\vh$.
In this way, the backbone representation determines how much of the low-level information should be propagated at a given denoising state. As shown in Table~\ref{tab:ablate_skip}, the adaGLU substantially improves generation performance over the standard identity gating function in both the symmetric and asymmetric skip topology settings.
It should be noted that the adaGLU block does not designate the nonlinear branch alone as the gate. Instead, its essential operation is the multiplicative interaction between the projected backbone and skip representations. Swapping the inputs changes only which representation undergoes the additional SiLU transformation, while preserving this cross-feature multiplicative modulation. We show experiment that compares $\text{adaGLU}(\vh, \vz)$ and $\text{adaGLU}(\vz, \vh)$ in the Appendix, Table~\ref{app_tab:swap_input}.

\subsection{Scale up model size}
We observe that the optimization bottlenecks are exposed when further increasing model size.
In Table~\ref{tab:model_init}, when we initially increase the depth from 12 to 20 layers, the nearly doubled parameters yield only a marginal improvement. This suggests that the deeper model is not effectively exploiting its additional capacity.
We attribute this degradation to the accumulation of activation variance in deep residual networks. A residual block can be written as $\vz_{l+1} = \vz_l + f_l(\vz_l), l\in[0,L-1]$, $L$ is model depth.
Assume the residual branch is uncorrelated with the main stream activations at initialization. Under conventional parameter initializations, the residual branch $f_l(\vz_l)$ contributes non-negligible accumulated variance, leading to $\text{Var}(\vz_l) + \text{Var}(f_l(\vz_l))$. Consequently, the activation magnitude can grow with depth, degrading performance in sufficiently deep networks. In gradient backpropagation, it can progressively amplify or attenuate different gradient directions. Thus, without controlling the scale of the residual branches, both the forward activations and backward gradients become increasingly poorly conditioned, making deeper models harder to optimize.

\begin{figure}
    \centering
    \includegraphics[width=\linewidth]{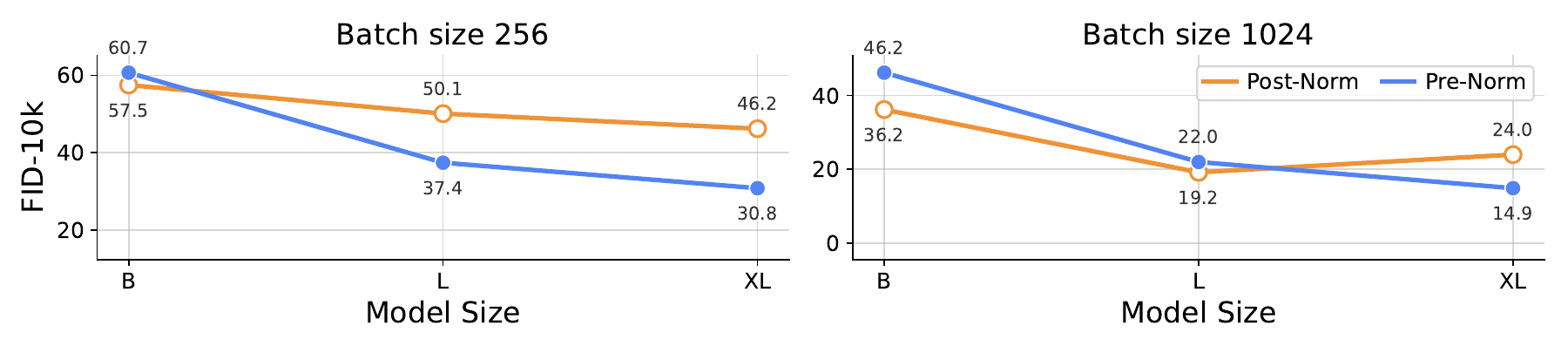}
    \caption{Comparing performance of Pre-Norm and Post-Norm adaGLU formulations across model sizes and training batch sizes. When trained with 1024 batch size, the Post-Norm adaGLU exhibits FID spikes at XL model size.}
    \label{fig:prenorm_postnorm}
\end{figure}

\begin{wraptable}{r}{.35\linewidth}
    \vspace{-12pt}
    \centering
    \caption{Address optimization bottleneck with proper weight init.}
    \label{tab:model_init}
    \begin{tabular}{l c}
        model & FID \\
        \toprule
        \rowcolor{jclightblue} 12-layer & 36.2 \\
        20-layer w/o init & 35.0 \\
        20-layer w init & \textbf{22.0}
    \end{tabular}
    \vspace{-10pt}
\end{wraptable}
To address this, we adopt a variance-preserving initialization strategy. Specifically, we follow standard practice~\citep{pmlr-v9-glorot10a, he2015delvingdeeprectifierssurpassing} and initialize weights of linear layers by independently sampling from a Gaussian distribution $\mathcal{N}(0, 1/d_{in})$, where $d_{in}$ is the input dimension of each layer. Then, we scale the output matrices of MLP and attention blocks by $\frac{1}{\sqrt{2*L}}$~\citep{marion2025scalingresnetslargedepthregime, zhang2019fixupinitializationresiduallearning}, where $L$ is the model depth. 
For each adaGLU block, we zero-initialize the output projection matrix ($\bm{W}_1$ in \eqref{eq:adaglu}).
We initialize the bias of all linear layers to zero.
Together, these choices bound the activation variance at initialization, ensuring that the forward pass can travel through an arbitrary number of initial random layers without magnitude explosion or decay. 
In early experiments, We find that scaling the weight matrices according to the depth of the corresponding encoder or decoder is slightly better than using the total model depth. 
We defer a formal analysis to the Appendix~\ref{app_sec:intialization_analysis}.

\begin{figure}[t]
    \centering
    \includegraphics[width=1.0\linewidth]{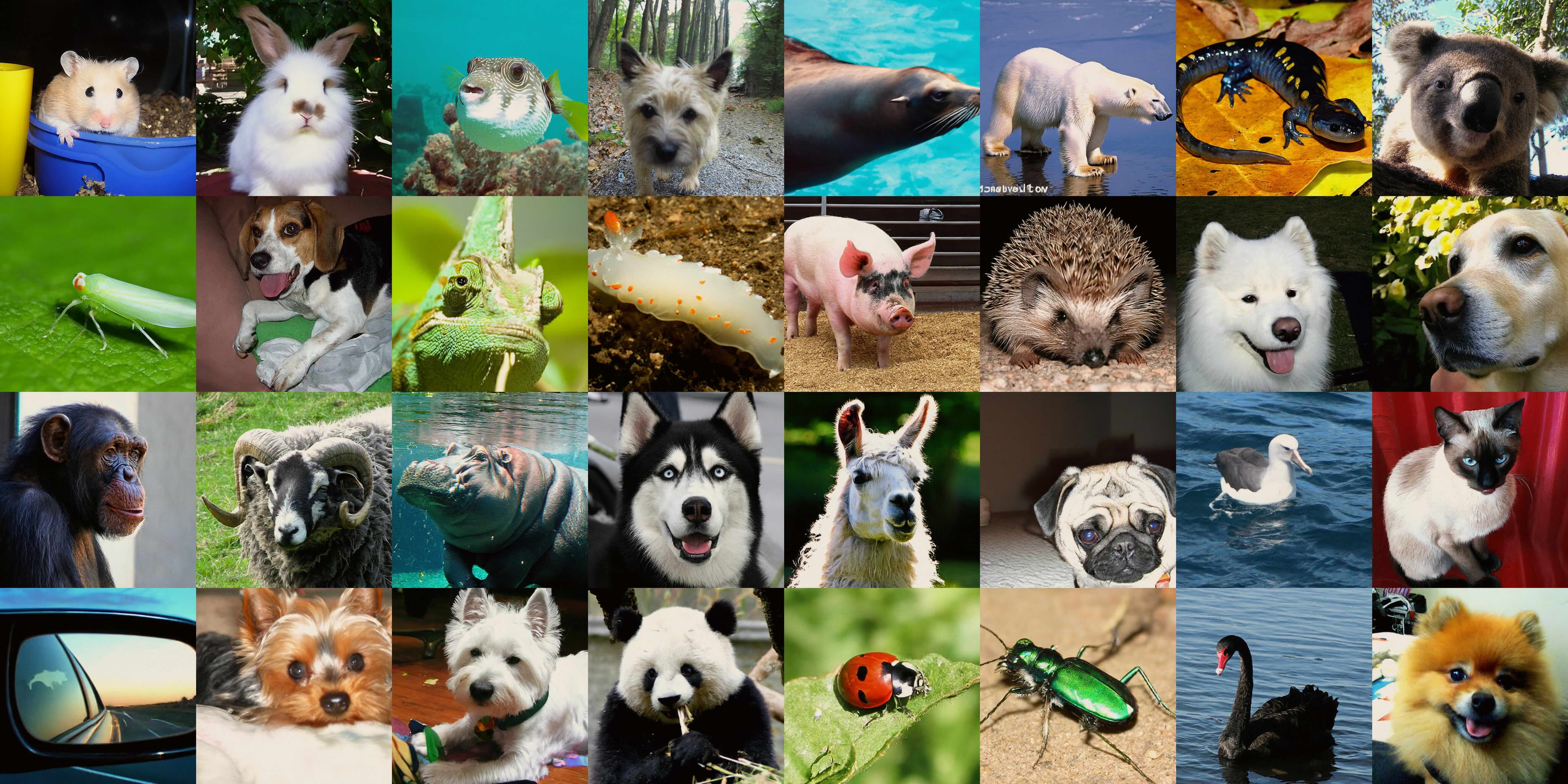}
    \caption{Quantitative results of class-to-image generation on ImageNet-512. We display more results in the Appendix.}
\end{figure}

\textbf{Stabilize training with Pre-Norm formulation.} 
Post-Norm architectures are known to become increasingly difficult to optimize at large model scales~\citep{liu2023understandingdifficultytrainingtransformers}, making optimization more sensitive to hyperparameter choices.
We observe a similar phenomenon in our experiments with the Post-Norm adaGLU defined in \eqref{eq:postnorm_skip}. 
In particular, when training a 30-layer and 1152-dimension ViT-XL model, both the loss and FID exhibit notable spikes (Appendix Figure~\ref{app_fig:loss_curve_prepostnorm_adaglu}).
We therefore introduce a Pre-Norm variant in which the normalization layer is applied to the backbone feature before input to adaGLU and replace the conventional fusion operator with simple addition. Specifically, we define
\begin{equation}
    \vz = \text{adaGLU}(\vh, \text{Norm}(\vz)) + \vz.
    \label{eq:prenorm_skip}
\end{equation}
We refer to this formulation as Pre-Norm adaGLU.
We compare it with the Post-Norm adaGLU defined in ~\eqref{eq:postnorm_skip} in Figure~\ref{fig:prenorm_postnorm} across various model sizes and training batch sizes.
The Pre-Norm formulation avoids the training instability (or FID spike ) observed for the Post-Norm formulation at large model sizes (XL size). Besides, its scaling behavior also shows better consistency across different training batch sizes.

\subsection{Discussion}
Our final architecture combines the two modifications identified above: asymmetric long skip connections and adaptive feature fusion with adaGLU. The resulting design retains a standard transformer backbone while using the long skip primarily as a pathway for low-level information.
Figures~\ref{fig:pit} and ~\ref{fig:adaglu_vs_swiglu} show the overall architecture and the Pre-Norm adaGLU, respectively. We also compare our final solution to prior U-Net variants in Table~\ref{tab:compare_skip_connection}.

We further examine two architectural choices. First, the input feature $\vh$ used by adaGLU is taken from the tokens immediately after patch embedding and before the first ViT layer. One alternative is to introduce a separate patch-embedding layer for each adaGLU block. Our experiments (Appendix Table~\ref{app_tab:share_pe}) on these two settings show only marginal differences in FID on ImageNet-256, so we use the shared input embedding for simplicity.
Second, we use an equal number of encoder and decoder layers. For example, a 12-layer model contains 6 encoder layers and 6 decoder layers. Although increasing the number of decoder layers monotonically improves performance in our ablations (Appendix Table~\ref{app_tab:depth}), we retain a uniform layer allocation throughout the experiments and find that it's sufficient for strong generation performance.

%% file: experiment.tex
\section{Experiment}
\label{sec:experiment}
\begin{table}[t]
    \centering
    \caption{Results on ImageNet-256.  We categorize the diffusion models trained with additional self-supervised learning (SSL) stages or using external semantic encoders as \textit{w SSL}. Following DiT, we report GFLOPs of our model as: $M\cdot N$, where $M$ denotes model parameters and $N$ is the token sequence length. For latent diffusions, we display the parameters of the VAE decoder and the diffusion model sequentially.}
    \label{tab:imagenet256}
    \footnotesize
    \begin{tabular}{l c c c c c c c c}
        method & FID$\downarrow$ & IS$\uparrow$ & pre. & rec. & epoch & \#params & GFLOPs \\
        \toprule
        \textit{w SSL} \\
        \midrule
        USP DiT-XL/2~\citep{chu2025usp} & 2.33 & 267.0 & - & - & 240 & 84 + 675M & 312 + 119  \\
        PixelGen~\citep{ma2026pixelgenimprovingpixeldiffusion} & 1.83 & 293.6 & 0.79 & 0.63 & 160 & 676M & - \\
        PixelU-H/16~\citep{guo2026pixeluushapedtransformerefficient} & 1.63 & 305.9 & 0.79 & 0.64 & 600 & 1.2B & 137 \\
        PixelDiT-XL~\citep{yu2026pixelditpixeldiffusiontransformers} & 1.61 & 292.7 & 0.78 & 0.64 & 320 & 797M & 311 \\
        EPG-G/16~\citep{lei2026vaeendtoendpixelspacegenerative} & 1.58 & 298.4 & 0.80 & 0.63 & 1600 & 1391M & 321 \\
        HyperDiT-H~\citep{he2026hyperdithyperconnectedtransformershighfidelity} & 1.56 & 306.5 & 0.80 & 0.64 & 600 & 952M & - \\
        REPA~\citep{yu2025representationalignmentgenerationtraining}  & 1.42 & 305.7 & 0.80 & 0.65  & 800 &  84 + 675M & 312 + 119\\
        RAE~\citep{zheng2025diffusiontransformersrepresentationautoencoders} & \textbf{1.13} & 262.9 & - & -  & 800  & 415 + 839M & 107 + 146\\
          \midrule
        \textit{train from scratch} \\
        \midrule
         LDM~\citep{rombach2022highresolutionimagesynthesislatent} & 3.60 & 247.7 & 0.87 & 0.48 & 167 & 55 + 400M & 336 + 104\\
         U-ViT-H/2~\citep{bao2023worthwordsvitbackbone} & 2.29 & 263.9 & 0.82 & 0.57 & 400 & 84 + 501M & 312 + 133 \\
         MaskDiT~\citep{Zheng2024MaskDiT} & 2.28 & 276.6 & 0.80 & 0.61 & 1600 & 84 + 675M & 312 + 119  \\
         DiT-3B-G~\citep{gao2024luminat2xtransformingtextmodality} & 2.10 & 304.4 &  0.82 & 0.60 & 340 & 3B & - \\
         DAR~\citep{xu2026rethinkingcrosslayerinformationrouting} & 2.05 &  270.1 & 0.82 & 0.60 & 400 & 740M & - \\
         JiT-G/16~\citep{li2025basicsletdenoisinggenerative} & 1.82 & 292.6 & 0.79 & 0.62 & 600 & 2B & 383 \\
         SiD2~\citep{hoogeboom2025simplerdiffusionsid215} & 1.72 & - & - &  - & - & - & 137 \\
        \rowcolor{jclightblue} \textbf{PiT-B/16} & 3.60 & 237.9 & 0.83 & 0.50  & 800 & 123M & 32  \\
        \rowcolor{jclightblue}\textbf{PiT-L/16}  & 2.24 & 259.8 & 0.81 & 0.59 & 800 & 339M & 87 \\
        \rowcolor{jclightblue}\textbf{PiT-XL/16} & 1.78 & 278.7 & 0.79 & 0.62 & 800 & 640M & 165 \\
        \rowcolor{jclightblue}\textbf{PiT-G/16} & \textbf{1.62} & 293.1 & 0.79 & 0.64 & 800 & 1.6B & 403 \\
        \bottomrule
    \end{tabular}

\end{table}

\begin{table}[t]
    \centering
    \caption{Results on ImageNet-512. The settings are the same as Table~\ref{tab:imagenet256}.}
    \label{tab:imagenet512}
    \footnotesize
    \begin{tabular}{l c c c c c c c}
        method & FID$\downarrow$ & IS$\uparrow$ & prec. & rec. & epoch  & \#params & Gflops \\
        \toprule
        EPG-L/32~\citep{lei2026vaeendtoendpixelspacegenerative} &  2.35 & 295.4 & 0.82 & 0.57 & 800 &  540M & 113 \\
        DeCo-XL/16~\citep{ma2026decofrequencydecoupledpixeldiffusion} & 2.22 & 290.0 & 0.80 & 0.60 & 340 & 682M & - \\
        \midrule
        U-ViT-H/2~\citep{bao2023worthwordsvitbackbone} & 4.05 & 263.8 & 0.84 & 0.48 & 400 & 84 + 501M & 1260 + 133 \\
        SiT-XL/2~\citep{ma2024sitexploringflowdiffusionbased} & 2.62 & 252.2 & 0.84 & 0.57 & 600 & 84 + 675M  & 1260 + 525 \\
        SiD2~\citep{hoogeboom2025simplerdiffusionsid215} & 2.19 & - & - & - & - & - & 137 \\
        PixelU-H/32 & 1.92 & 322.1 & 0.80 & 0.58 & 600 & 1.2B & - \\
        PixelDiT-XL & 1.81 & 278.6 & 0.78 & 0.67 & 850 & 797M & 1352 \\
        EDM2~\citep{Karras2024edm2} & 1.81 & - & - & - & - & 1.5B & 552 \\
        JiT-G/32~\citep{li2025basicsletdenoisinggenerative} & 1.78 & 306.8 & - & - & 600 & 2B & 384 \\
        \rowcolor{jclightblue}\textbf{PiT-B/32} & 4.16 & 256.8 & 0.84 & 0.46 & 800 & 126M & 33 \\
        \rowcolor{jclightblue}\textbf{PiT-L/32} & 2.64 & 267.7 & 0.83 & 0.54 & 800 & 343M & 88 \\
        \rowcolor{jclightblue}\textbf{PiT-XL/32} & 2.10 & 290.9 & 0.81 & 0.60 & 800 & 646M & 167 \\
        \rowcolor{jclightblue}\textbf{PiT-G/32} & \textbf{1.70} & 313.2 & 0.80 & 0.62 & 800 & 1.6B & 405 \\
        \bottomrule
    \end{tabular}

\end{table}

\textbf{Settings.} We use the same set of hyper-parameters across all experiments, without scale- or resolution-specific tuning.  The resulting performance is therefore obtained without optimizing the training configuration for each setting. 
For example, we use 2e-4 learning rate, Adam optimizer with betas equal (0.9, 0.95), and 0.9999 ema decay, 1024 batch size. For text-to-image generation, we reduce the batch size to 512 while keeping all other settings unchanged.
Besides, we uses patch size that linearly scales with input image resolution (e.g., 16 and 32 for resolutions 256 and 512, respectively), maintaining consistent computation cost across various resolutions. We defer more implementation details to the Appendix. 

\textbf{Training \& Evaluation.} For class-to-image generation, we train models on ImageNet-1K for 1000K steps (800 epochs).
We report the FID~\citep{NIPS2017_8a1d6947}, IS~\citep{salimans2016improvedtechniquestraininggans}, Precision and Recall~\citep{kynkäänniemi2019improvedprecisionrecallmetric} of models with 50K samples. We use reference statistics from ADM~\citep{dhariwal2021diffusionmodelsbeatgans}. 
For text-to-image generation, we use FLAN-T5-Large encoder~\citep{chung2022scalinginstructionfinetunedlanguagemodels} to encode prompt conditions.
Following the settings of MiniT2I~\citep{minit2i2026}, we first train the model on CC12M~\citep{changpinyo2021conceptual12mpushingwebscale} for 500K steps and then fine-tune it for 40K steps on 120K text-image pairs from three datasets, including Open-Dalle3\footnote{https://huggingface.co/datasets/OpenDatasets/dalle-3-dataset}, ShareGPT~\citep{chen2025sharegpt4oimagealigningmultimodalmodels}, and BLIP3o-60k~\citep{chen2025blip3ofamilyfullyopen}. We evaluate the model performance on GenEval and DPG-Bench and generate four images per test prompt without prompt rewriting.
We generate samples utilizing interval classifier-free guidance~\citep{ho2022classifierfreediffusionguidance, kynkaanniemi2024applying} and the Heun ODE sampler~\citep{karras2022elucidating} with 50 sampling steps. 

\subsection{Class-to-image generation}
We display our results on ImageNet-256 and ImageNet-512 in Table~\ref{tab:imagenet256} and Table~\ref{tab:imagenet512} respectively. When compared with prior methods, PiT reaches competitive performance. On ImageNet-256, it surpasses several recent methods trained with an additional training stage or external models, including USP and PixelGen, and is on par with PixelDiT and EPG.

\textbf{Comparing with models with U-Net-style skip connection}, such as U-ViT, PixelU, DAR, and EDM2, our models yield outstanding results.
Among them, EDM2 shows notable performance with its magnitude-preserving network designs. In comparison, our results reveal that such additional structural complexity is unnecessary for building a strong pixel denoiser.

\textbf{Scalability.} PiT maintains strong performance as the model size increases from the small-scale baseline to 1.6B parameters. 
This scaling behavior becomes stronger and more pronounced on ImageNet-512, due to the increased complexity of the generative modeling task.

\begin{table}[t]
        \centering
        \small
        \caption{Benchmark text-to-image generation performance on GenEval and DPG-Bench. We display parameters of the text encoder and diffusion model sequentially.}
        \label{tab:t2i}
        \begin{tabular}{l c c c}
             method & GenEval & DPG-Bench & \#params \\
             \toprule
              PixelDiT-T2I~\citep{yu2026pixelditpixeldiffusiontransformers} & 0.78 & 83.7 & 2.6+1.3B\\
              PixelGen-XXL/16~\citep{ma2026pixelgenimprovingpixeldiffusion} & 0.79 & - & 1.7+1.1B\\
              DeCo-XXL/16~\citep{ma2026decofrequencydecoupledpixeldiffusion} & \textbf{0.86} & 81.4 & 1.7+1.1B \\
              \textbf{PiT-L/32} & 0.82 & \textbf{84.3} & 341+344M \\
             \bottomrule
        \end{tabular}
\end{table}

\begin{figure}[t]
    \centering
    \includegraphics[width=1.0\linewidth]{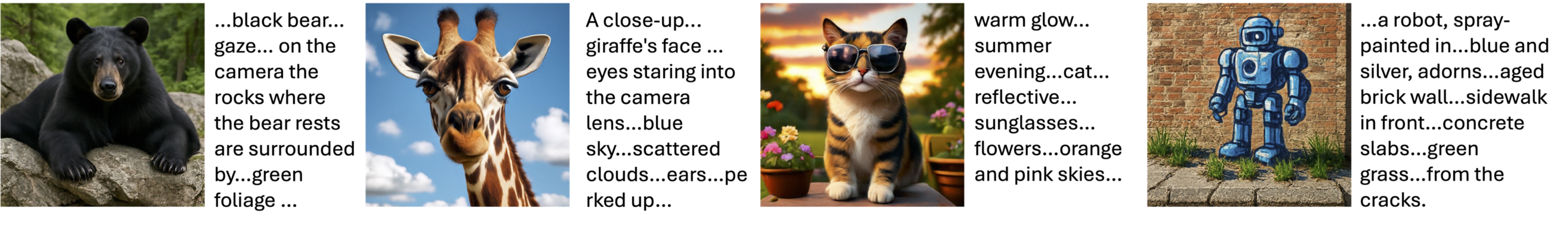}
    \caption{Qualitative results of text-to-image generation. More results are deferred to the Appendix.}
    \label{fig:t2i_qualitative}
\end{figure}

\subsection{Text-to-image Generation}
We show that our model natively supports text and image input without relying on cross attention or the two-stream architecture as in MMDiT~\citep{esser2024scalingrectifiedflowtransformers}.
Using a large patch size of 32, the input tokens to our model comprise three parts, including time condition (1), text embeddings (256), and image tokens (256), resulting in a total of 513 tokens. In Table~\ref{tab:t2i}, our middle-sized PiT-L/32 model reaches generation performance on par with prior works.

%% file: appendix.tex
\section{Experiment settings}
\label{app_sec:experiment_settings}

\textbf{Training.} We list the training and sampling hyper-parameters used in our experiments in Table~\ref{app_tab:appendix_hyper_params}.
As we adopt the variance-exploding diffusion model, to mitigate the model input variance, we scale the model input $\vx(t)$ with a coefficient $c_{in}=1/\sqrt{t^2+\sigma_{data}^2}$ following~\citet{karras2022elucidating}.

\textbf{Evaluation.} In class-conditional image generation experiments (Table~\ref{tab:imagenet256} and ~\ref{tab:imagenet512}), we generate 50 samples per class, aligned with the settings in JiT~\citep{li2025basicsletdenoisinggenerative}.

\textbf{Architecture.} 
We list our architectural settings in Table~\ref{app_tab:pit_architecture}. Note that for the 12-layer PiT-B in our main experiments, we adopt Post-Norm adaGLU as we observe Pre-Norm adaGLU prefers deeper network. We show this in Table~\ref{tab:pit_b_vs_deep}.

\begin{table}[h]
    \centering
    \caption{The architecture settings of PiTs.}
    \tablestyle{1.2pt}{1.1}
    \begin{tabular}{l c c c c c}
        \toprule
        model & layers & dim & mlp dim & adaGLU dim & heads  \\
        \midrule
        PiT-B & 12 & 768 & 2048 & 2048 & 12 \\
        PiT-Deep & 20 & 640 & 1707 & 1707 & 10 \\
        PiT-L & 20 & 1152 & 3072 & 3072 & 16  \\
        PiT-XL & 30 & 1152 & 3072 & 3072 & 16 \\
        PiT-G & 30 & 1536 & 6144 & 6144 & 16 \\
    \end{tabular}
    \label{app_tab:pit_architecture}
\end{table}

\begin{table}[h]
    \caption{Training \& sampling hyper-parameters in main experiments including class-to-image and text-to-image generation.}
    \label{app_tab:appendix_hyper_params}
    \centering
    \begin{tabular}{r c}
        hyper-parameter & value \\
        \toprule
         \rowcolor{jclightblue} \textbf{\textit{training}} & \\
         data augmentation & horizontal flip \\
         mixed precision & bf16 \\
         batch size  &  1024 \\
         learning rate & 2e-4 \\
         optimizer & Adam, $\beta_1,\beta_2$ = 0.9, 0.95 \\
         patch size & $\text{image\_size} / 16$ \\
         diffusion setting & $\sigma_{min}, \sigma_{max}, \sigma_{data}$ = 0.002, 80, 0.5 \\
         noise scale & $\text{image\_size} / 64$ \\
         weighting function & $\frac{1}{t_n - t_{n-1}}, t_n \in [\sigma_{min}, \sigma_{max}], N=1280$ \\
         time sampler & $\log(t)\sim \mathcal{N}(\mu, \delta^2)$, $\mu=-1.2$, $\delta=1.6$ \\
         ema decay & 0.9999 \\
         cfg drop & 0.1 \\
         \midrule
         \rowcolor{jclightblue} \textbf{\textit{sampling}} & \\
         sampler & Heun \\
         sampling step & 50 \\
         cfg interval & $[0.19, 1.61]$ (if used)\\
        \bottomrule
    \end{tabular}

\end{table}

\section{Apple-to-Apple Comparison with JiT}
\label{app+sec:a2a_jit}
Following the evaluation settings of JiT, we report FID scores of PiT models measured with the statistics of the training data, in Table~\ref{app_tab:a2a_jit}.
Note that the FID scores are slightly different from the numbers reported in our main experiments, where we use the statistics provided by ADM. 

On both datasets, utilizing the same training recipe, our PiT models consistently benefit from longer training time and larger model size.
On ImageNet-256, our model shows stronger scalability without the risk of overfitting. 
On ImageNet-512, with increased complexity, the number of model parameters becomes more essential for final generation performance. However, our 1.6B PiT-G/32 model surpasses JiT-G of 2B parameters. On both datasets, our models require neither additional in-context embeddings nor extensive hyper-parameter tuning, exhibiting clean scaling behavior. In Table~\ref{app_tab:jit_wo_incontext}, we evaluate a JiT-B model trained without in-context embeddings. The results reveal JiT's heavy reliance on in-context embeddings.

\begin{table}[]
    \centering
    \begin{minipage}[t]{.55\linewidth}
        \caption{Apple-to-apple comparison with JiT across various model sizes. We report model parameters from ImageNet-512 experiments.}
        \label{app_tab:a2a_jit}
        \begin{tabular}{l c c c c}
            model &  \#params & IN256 & IN512 & epoch \\
            \toprule
            JiT-L & 462M & 2.36 & 2.53 & 600 \\
            PiT-L & 343M & 2.40  & 2.92 & 600 \\
                  &      & 2.23 & 2.68 & 800 \\
            \midrule
            JiT-H  & 956M & 1.86 & 1.94  & 600 \\
            PiT-XL & 646M & 1.89 & 2.31  & 600 \\
                   &      & 1.76 & 2.10 & 800 \\
            \midrule
            JiT-G & 2B & 1.82 & 1.78 & 600 \\
            PiT-G & 1.6B & 1.61 & 1.83 & 600 \\
                  &      & 1.56 & 1.72 & 800
        \end{tabular}
    \end{minipage}
    \hfill
    \begin{minipage}[t]{.43\linewidth}
        \begin{minipage}[t]{\linewidth}
        \caption{Performance of JiT-B model without in-context embeddings.}
        \label{app_tab:jit_wo_incontext}
        \begin{tabular}{l c c}
            model & \#params & FID \\
            \toprule
            JiT & 131M & 38.1 \\
            JiT w/o in-context emb. & 131M & 47.1 \\
            PiT-B & 123M & 36.2 \\
            \\
        \end{tabular}
    \end{minipage}
    \begin{minipage}[t]{\linewidth}
            \caption{Comparing Post-Norm PiT-B with Pre-Norm PiT-Deep.}
            \label{tab:pit_b_vs_deep}
            \begin{tabular}{c c c c}
                \toprule
                model & adaGLU & \#params & FID \\
                \midrule
                PiT-B & Post-Norm & 123M & 3.60 \\
                PiT-Deep & Pre-Norm & 133M & 3.61 \\
            \end{tabular}
    \end{minipage}
    \end{minipage}
\end{table}

\section{Additional Experiment Results}
By default, we run experiments with PiT-XL/16 on ImageNet-256. We report FID measured with the statistics of ImageNet-256 training data. We additionally present six experiments, including:
\begin{enumerate}
    \item Ablation on decoder depth.
    \item Ablation on position of adaGLU in each transformer layer.
    \item Ablation on input embeddings: whether sharing the input patch embedding layer or using independent ones in each adaGLU block.
    \item Compare $\text{adaGLU}(\vh, \vz)$ with $\text{adaGLU}(\vz, \vh)$.
    \item Train with or without adaLN.
    \item Train with representation alignment loss~\citep{yu2025representationalignmentgenerationtraining}.
\end{enumerate}

In the ablation studies, we observe that increasing decoder depth monotonically improves performance (Table~\ref{app_tab:depth}). We show that rearranging the position of the adaGLU block (Table~\ref{app_tab:rearrange_block}),  utilizing input embeddings extracted from independent patch embedding layers (Table~\ref{app_tab:share_pe}), adopting $\text{adaGLU}(\vz, \vh)$ (Table~\ref{app_tab:swap_input}), or training with adaLN~\ref{app_tab:w_adaln} all produce similar results as the default settings.
When trained with representation alignment, the FID in Table~\ref{app_tab:wrepa} noticeably drops from 1.78 to 1.64.

\section{Interpretation of adaGLU behavior across noise levels}
\begin{figure}
    \centering
    \includegraphics[width=0.6\linewidth]{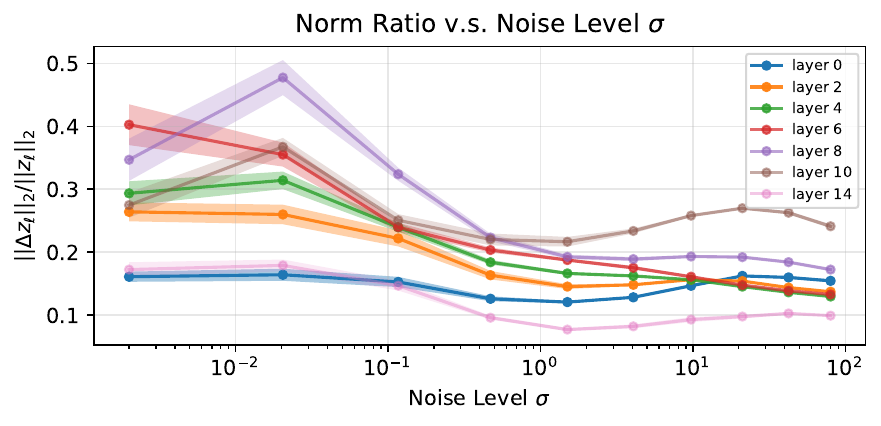}
    \caption{Interpretation of adaGLU behavior. 
    We visualize the ratio $\frac{||\Delta\vz_{\ell}||_2}{||\vz_{\ell}||_2}$ across various noise levels and decoder layers $\ell$. Here, $\vh$ denotes the adaGLU update $\Delta\mathbf{z}_{\ell}=\mathrm{adaGLU}(\mathbf{h}_{\ell},\mathbf{z}_{\ell})$, $\vz_{\ell}$ denotes the backbone feature before fusion. Checkpoint: PiT-XL at 800 training epochs.}
    \label{app_fig:skip_norm_ratio}
\end{figure}

For decoder layer $\ell$, let $\mathbf{z}_{\ell}$ denote the backbone feature before fusion and $\Delta\mathbf{z}_{\ell}=\mathrm{adaGLU}(\mathbf{h}_{\ell},\mathbf{z}_{\ell})$ denote the adaGLU update. We measure the relative update magnitude
$$
r_{\ell}(\sigma)=
\frac{\|\Delta\mathbf{z}_{\ell}\|_2}
{\|\mathbf{z}_{\ell}\|_2},
$$
averaged over 1000 samples at each noise level $\sigma$. The results in Figure~\ref{app_fig:skip_norm_ratio} show that $r_{\ell}(\sigma)$ decreases as the noise level increases across the examined decoder layers.

Since the features are fused by addition as $\mathbf{z}_{\ell}^{\mathrm{out}}=\mathbf{z}_{\ell}+\Delta\mathbf{z}_{\ell}$, this ratio measures the relative change that the adaGLU update induces in the backbone representation. Its decrease at higher noise levels therefore shows that the relative contribution of the adaGLU update diminishes. This observation is consistent with the network relying less on skip-path information when the input is heavily corrupted.

Conversely, the larger relative updates at low noise levels support our hypothesis that adaGLU helps propagate fine-grained input information to deeper layers to facilitate pixel prediction (Section 4.2, lines 249–254).

Besides, $\Delta\mathbf{z}_{\ell}$ contributes most in the intermediate decoder layers, showing that the adaGLU is deeply involved in the optimization of the network rather than acting as simple identity mapping.

\begin{table}
    \centering
    \caption{We ablate several design choices, including decoder depth, block sequence in decoder layer, and input embedding used in adaGLU. We report FID with 10K generated samples. The default settings we used in main experiments are highlighted in \hlc[jclightblue]{blue color}.}
    \subfloat[
        encoder-decoder depth.
        \label{app_tab:depth}
    ]{
        \begin{minipage}[t]{.42\linewidth}
            \centering
            \begin{tabular}{c c c c c}
                  enc & dec & FID & \#Params & GFLOPs \\
                  \toprule
                  \rowcolor{jclightblue}15 & 15 & 30.8 & 640M & 165 \\
                  24 & 6 & 33.2 & 545M & 141 \\
                  6 & 24 & 29.1 & 736M & 190 \\
            \end{tabular}
        \end{minipage}
        \hfill
    }\subfloat[
        block sequence.
        \label{app_tab:rearrange_block}
    ]{
    \begin{minipage}[t]{.32\linewidth}
        \centering
        \begin{tabular}{c c}
               & FID \\
              \toprule
             \rowcolor{jclightblue}adaGLU-attn-mlp & 30.8 \\
             attn-adaGLU-mlp & 30.6 \\
             \\
        \end{tabular}
    \end{minipage}
    \hfill
    }\subfloat[
     input embeddings.
    \label{app_tab:share_pe}
    ]{
    \begin{minipage}[t]{.25\linewidth}
        \centering
         \begin{tabular}{c c}
               & FID \\
              \toprule
             \rowcolor{jclightblue}shared & 15.3 \\
             independent & 15.8 \\
        \end{tabular}
    \end{minipage}
    }
    \\
    \vspace{1em}
    \subfloat[
        $\text{adaGLU}(\vh, \vz)$ v.s. $\text{adaGLU}(\vz, \vh)$.
        \label{app_tab:swap_input}
    ]{
        \begin{minipage}[t]{.42\linewidth}
            \centering
            \begin{tabular}{c c}
                   & FID \\
                  \toprule
                 \rowcolor{jclightblue} $\text{adaGLU}(\vh, \vz)$ & 30.8 \\
                 $\text{adaGLU}(\vz, \vh)$ & 30.0 \\
            \end{tabular}
        \end{minipage}
        \hfill
    }
    \subfloat[
        using adaLN.
        \label{app_tab:w_adaln}
    ]{
        \begin{minipage}[t]{.42\linewidth}
            \centering
            \begin{tabular}{c c}
                   & FID \\
                  \toprule
                 \rowcolor{jclightblue} PiT-XL & 15.3 \\
                 PiT-XL w/ adaLN & 15.0 \\
            \end{tabular}
        \end{minipage}
        \hfill
    }
\end{table}

\begin{table}
    \centering
    \caption{Training with representation alignment loss. We compare the performance of two PiT-XL/16 models, one is trained from scratch trained and the other is trained with representation alignment loss. We use default training settings of REPA and align the features of the $10$-th layer with the output of DINOv3-B model~\citep{siméoni2025dinov3}.}
    \label{app_tab:wrepa}
     \begin{tabular}{c c c c c c c}
           & FID & sFID & IS & pre. & rec. & epoch \\
          \toprule
         \rowcolor{jclightblue}from scratch & 1.78 & 4.35 & 278.7 & 0.79 & 0.62 & 800 \\
         w REPA & 1.64 & 4.30 & 284.8 & 0.80 & 0.62 & 800 \\
    \end{tabular}
\end{table}

\begin{figure}
    \centering
    \begin{minipage}[t]{.49\linewidth}
        \includegraphics[width=\linewidth]{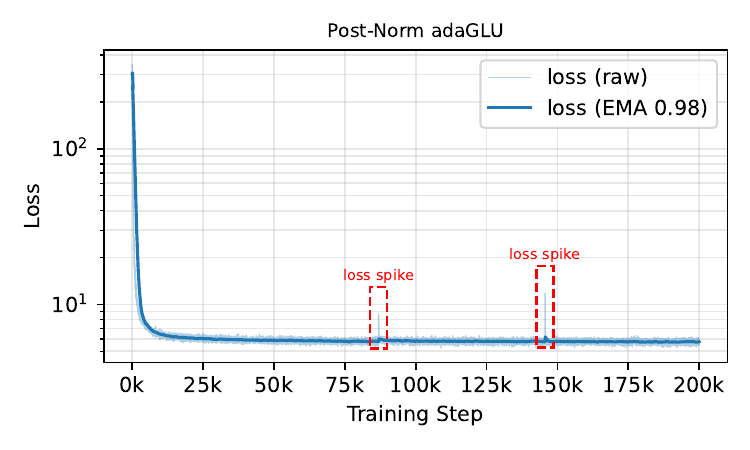}
    \end{minipage}\hfill
    \begin{minipage}[t]{.49\linewidth}
        \includegraphics[width=\linewidth]{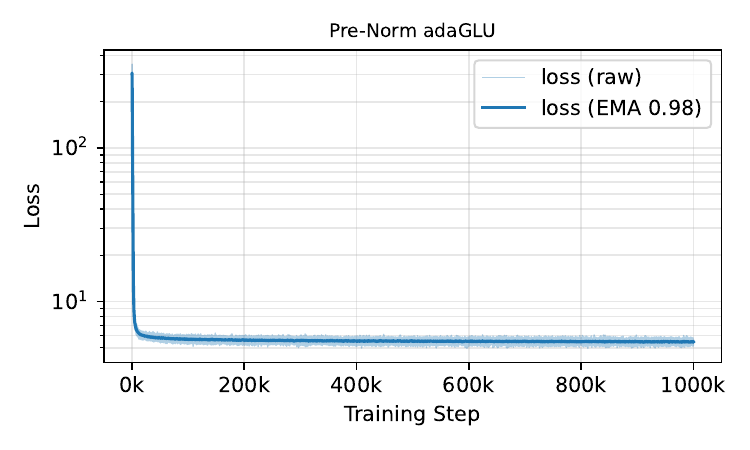}
    \end{minipage}
    \caption{Loss curve of Pre-Norm (right) and Post-Norm (left) adaGLU. We observe a notable loss spike when training with Post-Norm adaGLU. In contrast, the training of Pre-Norm adaGLU is stable.}
    \label{app_fig:loss_curve_prepostnorm_adaglu}
\end{figure}

\section{Does our design benefit latent-space models}

We evaluate asymmetric skips and adaGLU on latent-space U-ViT-B. The results are shown in~\ref{app_tab:latent_space}. Replacing the U-Net-style skips with asymmetric skips leaves performance essentially unchanged, whereas adding adaGLU improves FID-10K to 62.0. Thus, the benefit of adaGLU is consistent with our pixel-space findings, while the effect of asymmetric skips differs between the two experiments.  These preliminary results suggest that adaGLU’s multiplicative feature fusion is not inherently pixel-specific and can also benefit latent-space diffusion. In contrast, the benefit of asymmetric input routing may depend more on the input representation. 

\begin{table}[]
    \centering
    \begin{tabular}{l c}
        model & FID-10K \\
        \toprule
        U-ViT-B & 68.7 \\
        U-ViT-B + asym skip & 69.0 \\
        U-ViT-B + adaGLU & \textbf{62.0} \\
        \bottomrule
    \end{tabular}
    \caption{
        Evaluating asymmetric skips and adaGLU on latent-space U-ViT-B. We use sd-vae-ft-ema\textsuperscript{$\dagger$} as the VAE. All latent-space variants use the same pixel-derived hyper-parameters without latent-specific tuning. These experiments therefore provide a controlled architectural comparison rather than an optimized latent-diffusion results. $^\dagger$: \url{https://huggingface.co/stabilityai/sd-vae-ft-ema}.
    }
    \label{app_tab:latent_space}
\end{table}

\section{Model initialization}
\label{app_sec:intialization_analysis}
Let $\vx_l$ denote the input tensor to the $l$-th transformer layer, which consists of an attention block $f_1(\cdot)$ and an MLP block $f_2(\cdot)$. The forward pass of a Pre-LN block is defined as:
\begin{align*}
    \vh_l &= \vx_l + f_1(\text{LN}(\vx_l)) \\
    \vx_{l+1} &= \vh_l +f_2(\text{LN}(\vh_l))
\end{align*}

We evaluate the variance $\bm{V}_l = \text{Var}(\vx_l)$ sequentially through the network, assuming the residual branches $f_1(\text{LN}(\vx_l)), f_2(\text{LN}(\vh_l))$ are uncorrelated with the main stream activations $\vx_l$ at initialization. We assume the model input has a zero mean. 
The derivation is not a rigorous mathematical analysis but an intuitive explanation of the underlying rationale of our initialization strategy.

\textbf{Variance after the attention block}. The LayerNorm operation standardizes its input such that $\text{Var}(\text{LN}(\vx_l)) = 1$. Under He initialization (fan-in mode), the weight matrices are scaled to preserve the variance of normalized inputs through the linear layers. Let the variance of the unscaled attention output be a constant $\text{Var}(f_1(\text{LN}(\vx_l)))=C_A\approx1$. By scaling the output matrix by $\frac{1}{\sqrt{2L}}$, the variance of the attention branch is 
\begin{align*}
\text{Var}\left( \frac{1}{\sqrt{2L}} f_1(\text{LN}(x_l)) \right) &= \left( \frac{1}{\sqrt{2L}} \right)^2 
\text{Var}(f_1(\text{LN}(x_l))) = \frac{C_A}{2L}    
\end{align*}
The variance of the intermediate representation $h_l$ is the sum of the main stream variance and the branch variance:
\begin{align*}
    \text{Var}(h_l) &= \text{Var}(x_l) + \frac{C_A}{2L}\\ 
    &= V_l + \frac{C_A}{2L}
\end{align*}
\textbf{Variance after the MLP Block.} Similarly, the intermediate tensor $h_l$ is normalized before entering the MLP, yielding $\text{Var}(\text{LN}(h_l)) = 1$. He initialization preserves this unit variance through the hidden layers, producing an unscaled output variance $C_M \approx 1$. Scaling the MLP output matrix by $\frac{1}{\sqrt{2L}}$ results in a branch variance of $\frac{C_M}{2L}$. Adding this to the intermediate main stream gives the variance at the output of the $l$-th layer:
\begin{align*}
V_{l+1} &= \text{Var}(h_l) + \frac{C_M}{2L} \\
&= V_l + \frac{C_A}{2L} + \frac{C_M}{2L}  \\
&= V_l + \frac{C_A + C_M}{2L} \\
\end{align*}
\textbf{Variance after L layers.} To find the final variance $V_L$ after passing through all $L$ layers, we unroll the recurrence relation from the initial embedding $x_0$:
\begin{align*}
    V_L &= V_0 + \sum_{l=1}^{L} \frac{C_A + C_M}{2L} \\
    &= V_0 + L \left( \frac{C_A + C_M}{2L} \right) \\
    &= V_0 + \frac{C_A + C_M}{2}
\end{align*}
Without the $\frac{1}{\sqrt{2L}}$ factor, the branch variances $C_A$ and $C_M$ would be added directly to the main stream at each layer, resulting in $V_L = V_0 + L(C_A + C_M)$. The variance would scale linearly with depth $O(L)$. 
By scaling the output matrices by $\frac{1}{\sqrt{2L}}$, the depth factor $L$ is analytically canceled out in the sum. The final variance $V_L$ becomes strictly bounded by a constant $\mathcal{O}(1)$, regardless of how deep the network is scaled. When adding an adaGLU block to decoder layers, we zero-initialize its output matrix and use standard He initialization for the other linear layers. This ensures our prior analysis remains valid.

\section{Additional Qualitative Results}
\begin{figure}
    \centering
    \begin{minipage}[t]{\linewidth}
        \centering
        \begin{minipage}[t]{.48\linewidth}
            \includegraphics[width=\linewidth]{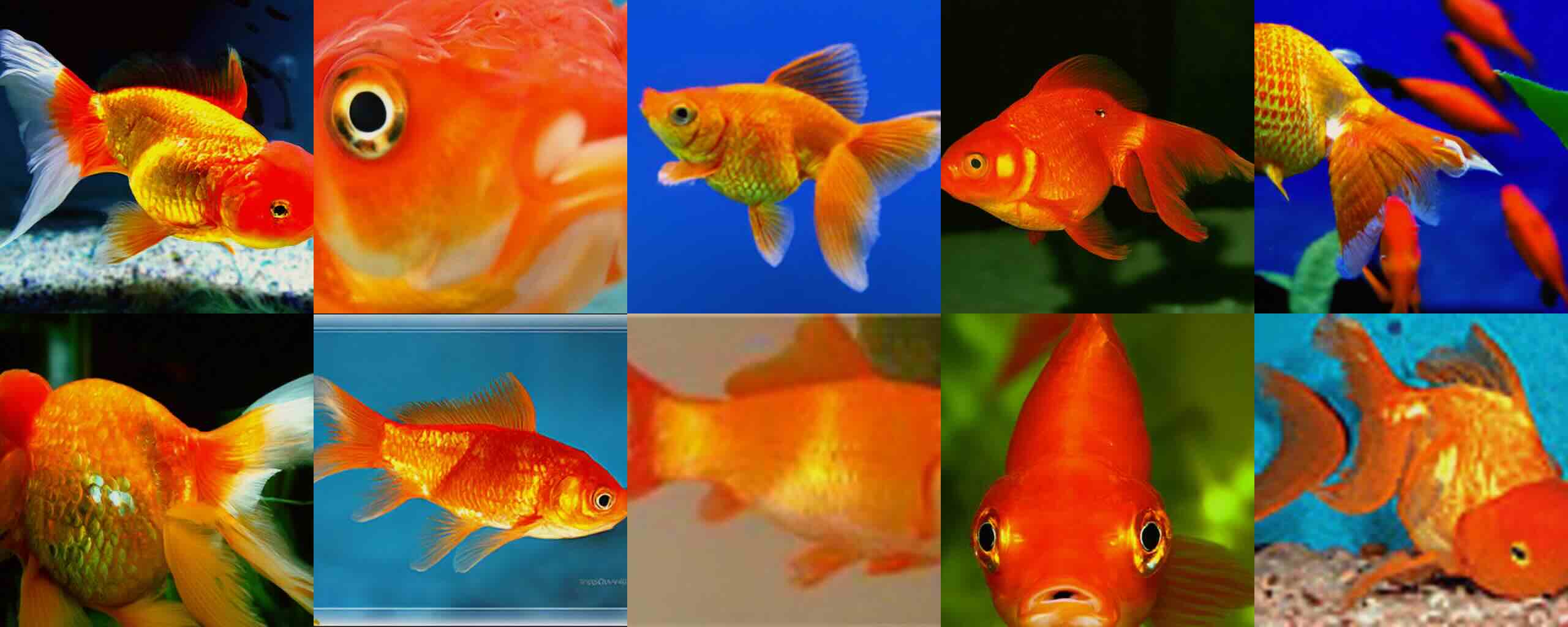}
        \end{minipage}
        \begin{minipage}[t]{.48\linewidth}
            \includegraphics[width=\linewidth]{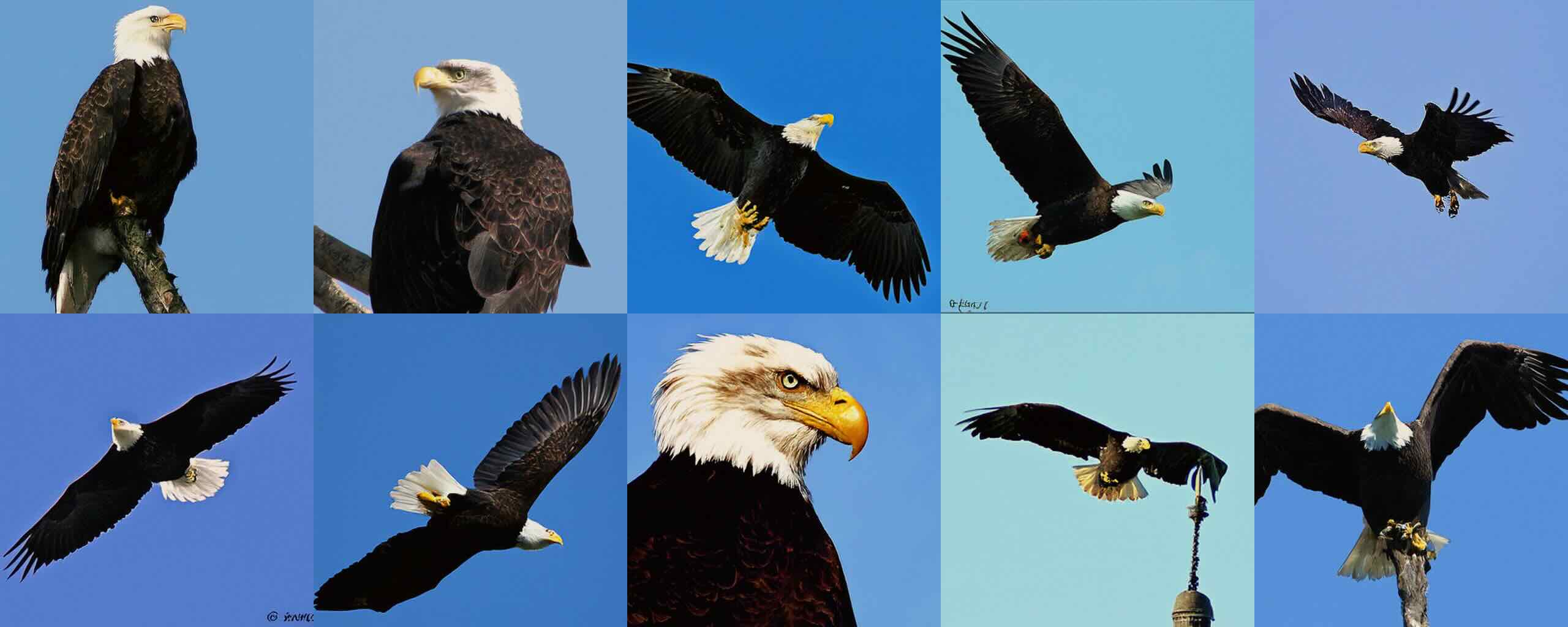}
        \end{minipage}
    \end{minipage}
    \begin{minipage}[t]{\linewidth}
        \centering
        \begin{minipage}[t]{.48\linewidth}
            \includegraphics[width=\linewidth]{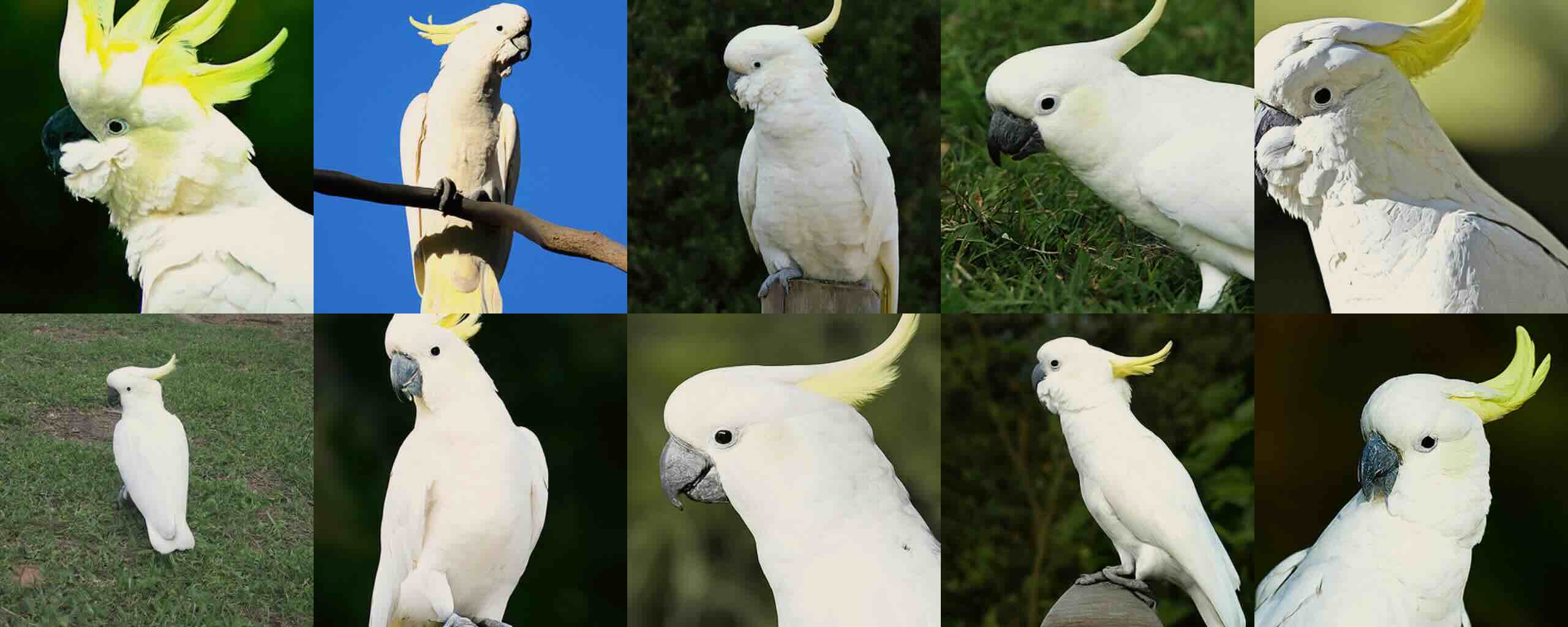}
        \end{minipage}
        \begin{minipage}[t]{.48\linewidth}
            \includegraphics[width=\linewidth]{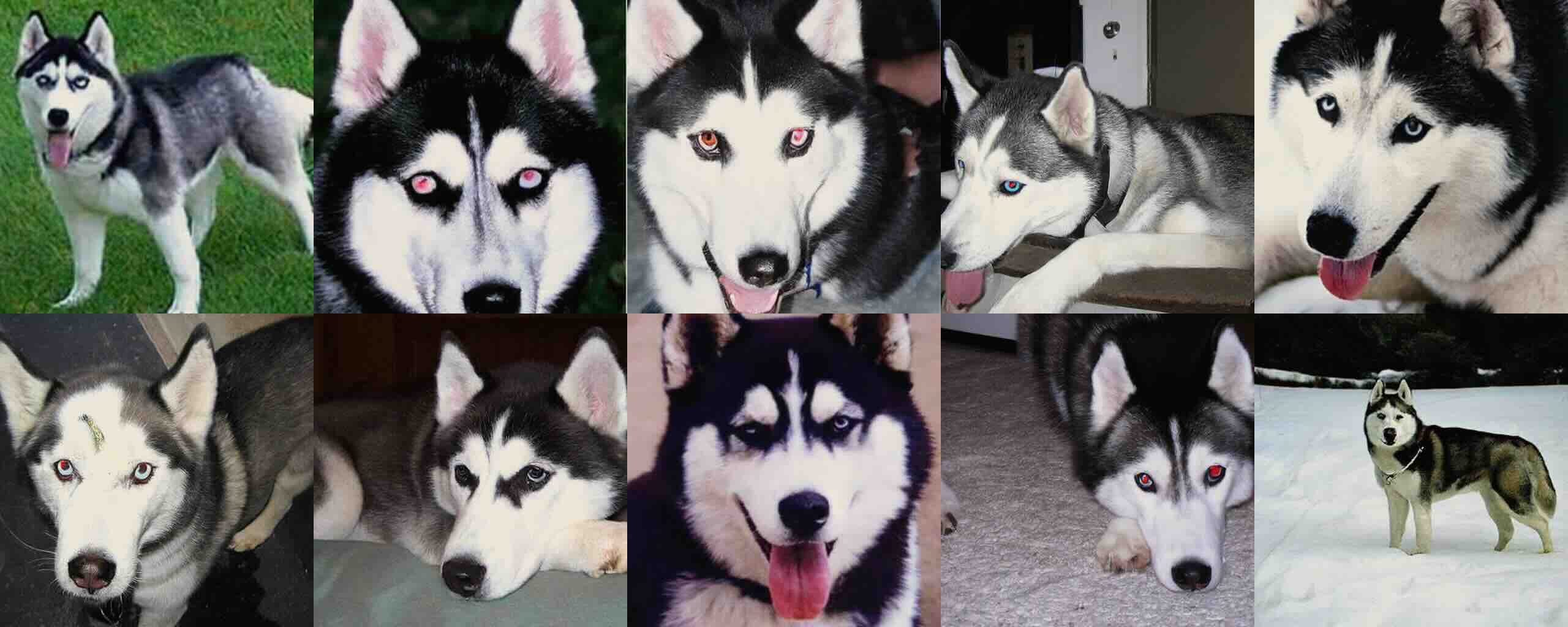}
        \end{minipage}
    \end{minipage}
    \begin{minipage}[t]{\linewidth}
        \centering
        \begin{minipage}[t]{.48\linewidth}
            \includegraphics[width=\linewidth]{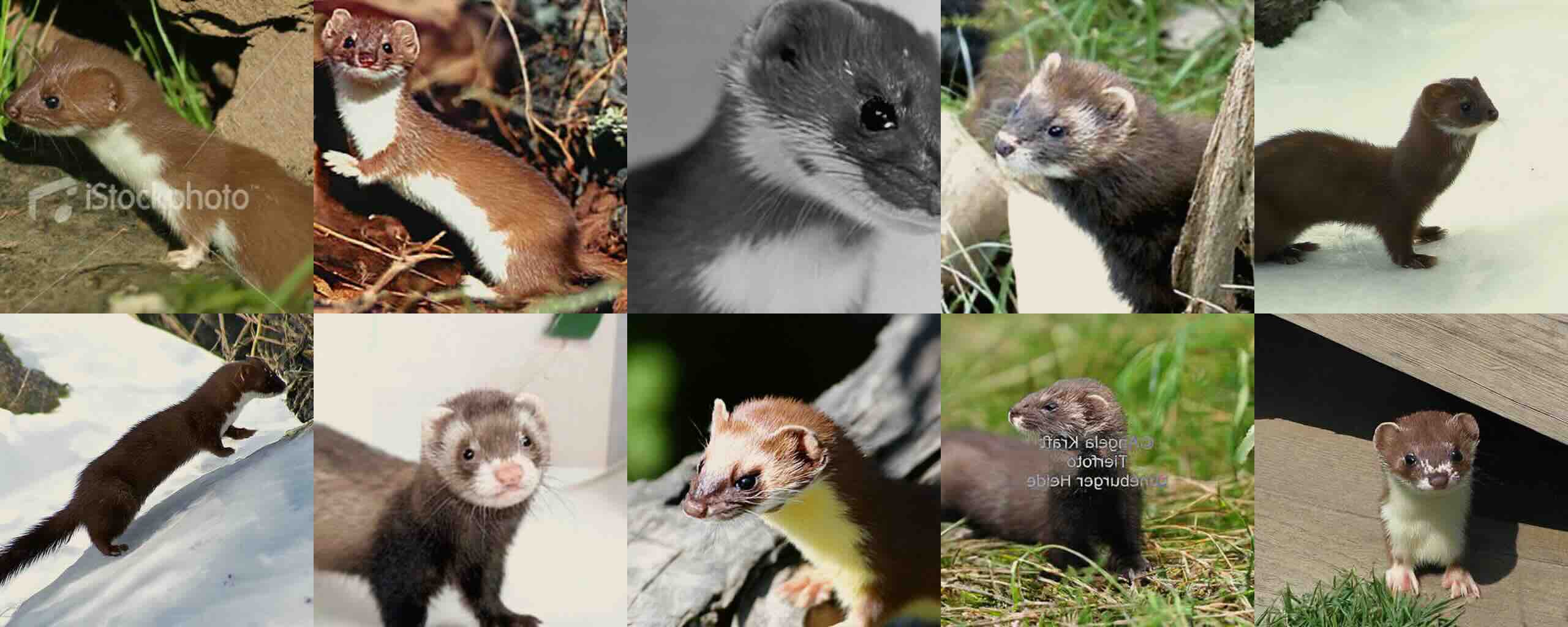}
        \end{minipage}
        \begin{minipage}[t]{.48\linewidth}
            \includegraphics[width=\linewidth]{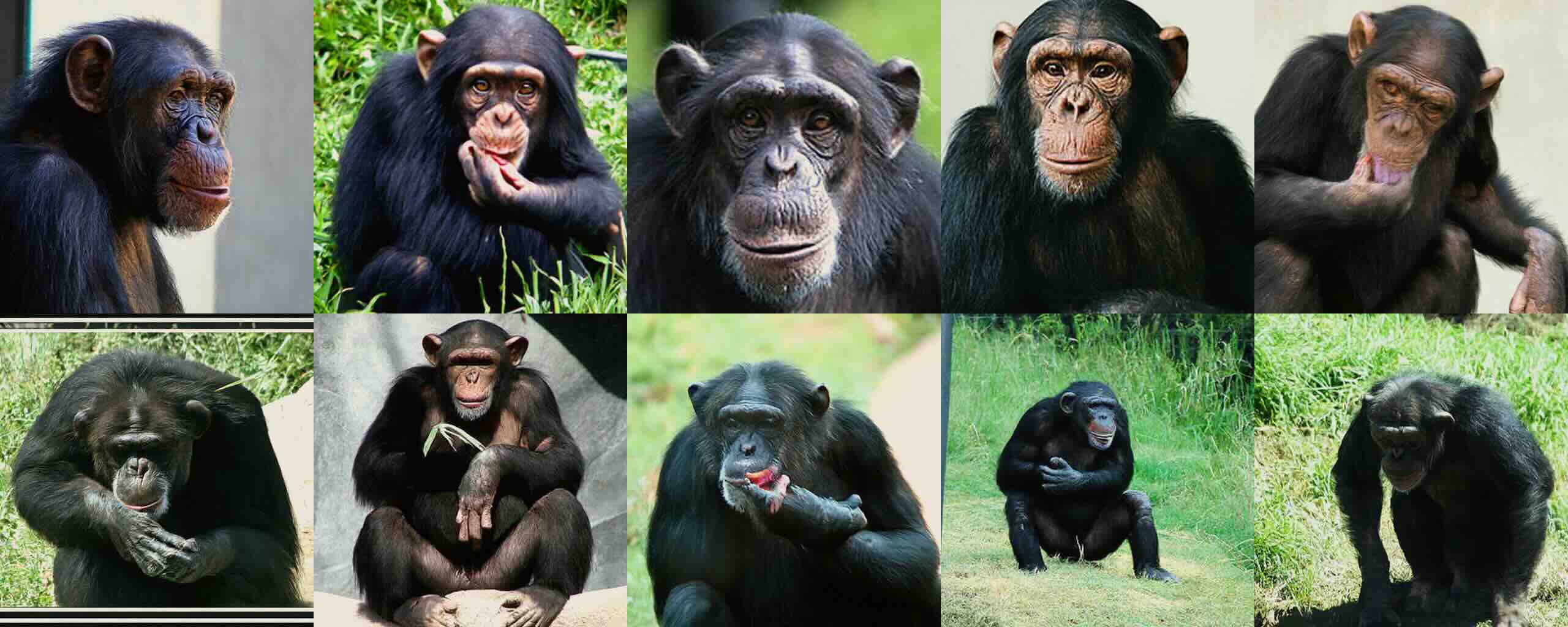}
        \end{minipage}
    \end{minipage}
    \begin{minipage}[t]{\linewidth}
        \centering
        \begin{minipage}[t]{.48\linewidth}
            \includegraphics[width=\linewidth]{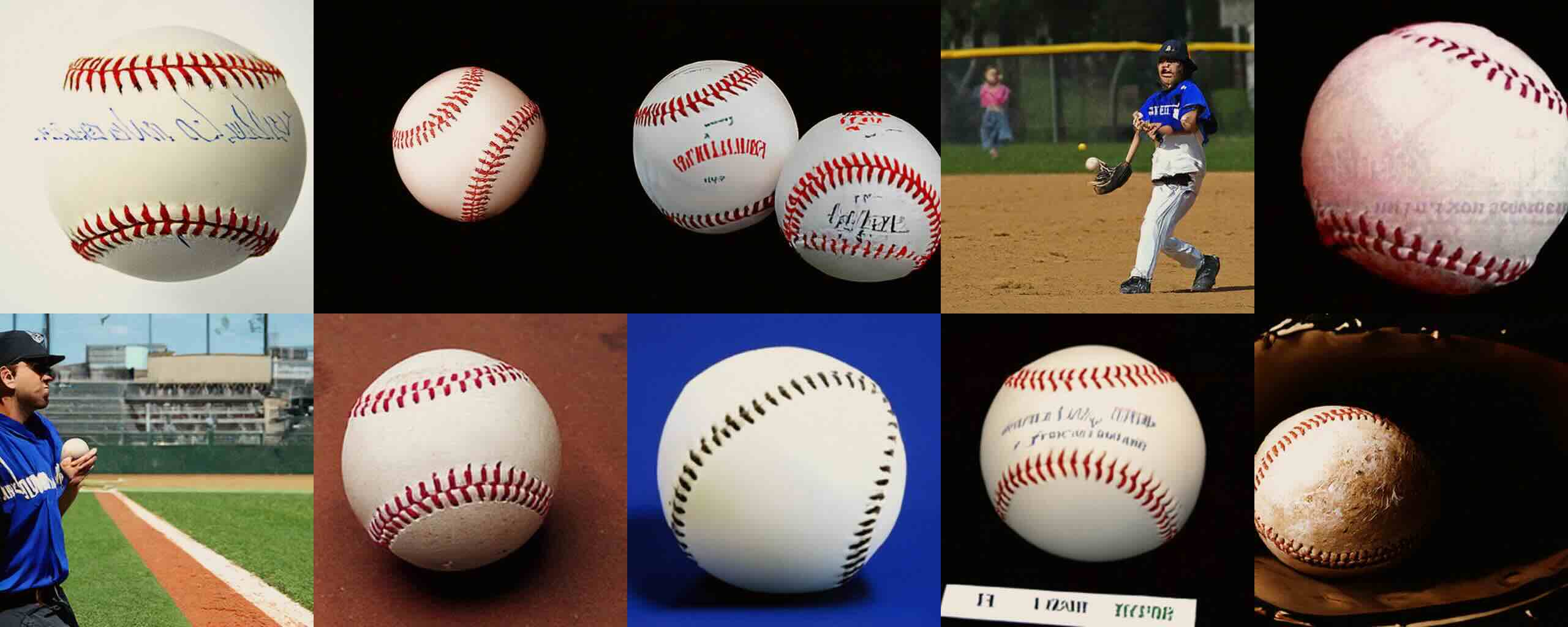}
        \end{minipage}
        \begin{minipage}[t]{.48\linewidth}
            \includegraphics[width=\linewidth]{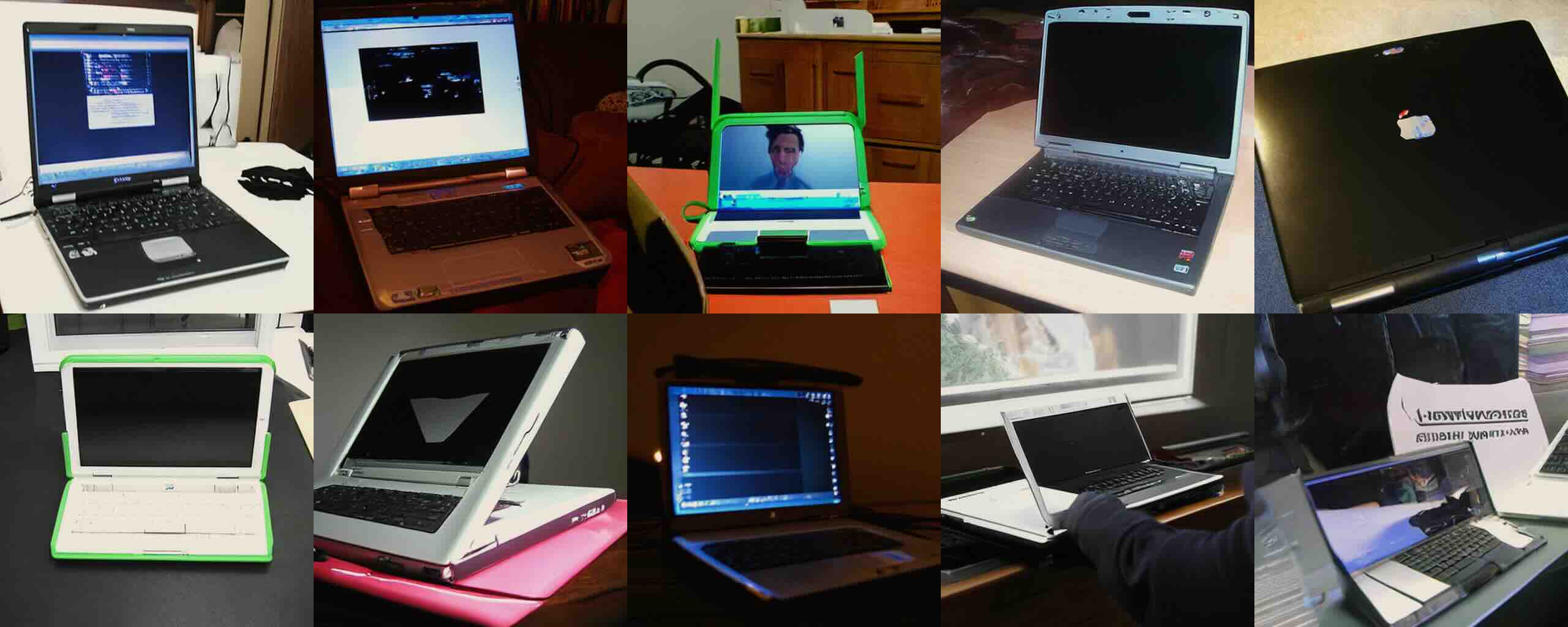}
        \end{minipage}
    \end{minipage}
    \begin{minipage}[t]{\linewidth}
        \centering
        \begin{minipage}[t]{.48\linewidth}
            \includegraphics[width=\linewidth]{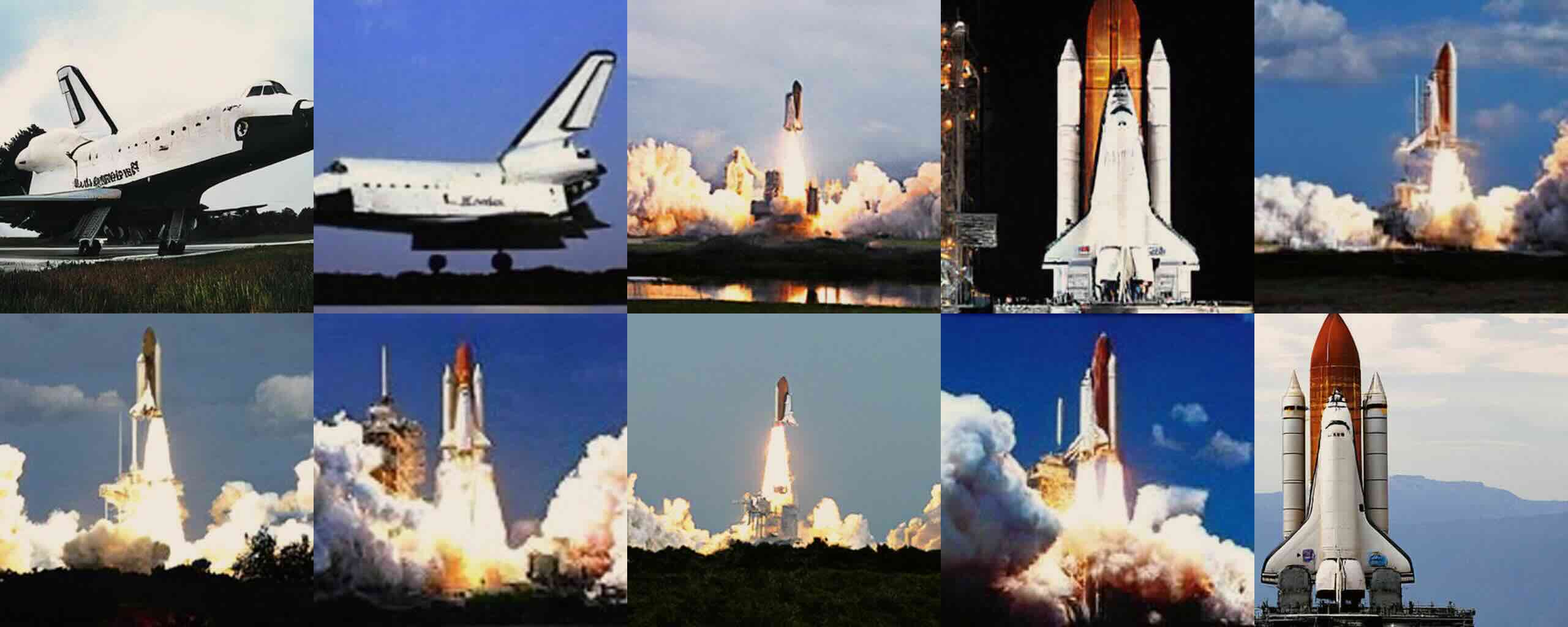}
        \end{minipage}
        \begin{minipage}[t]{.48\linewidth}
            \includegraphics[width=\linewidth]{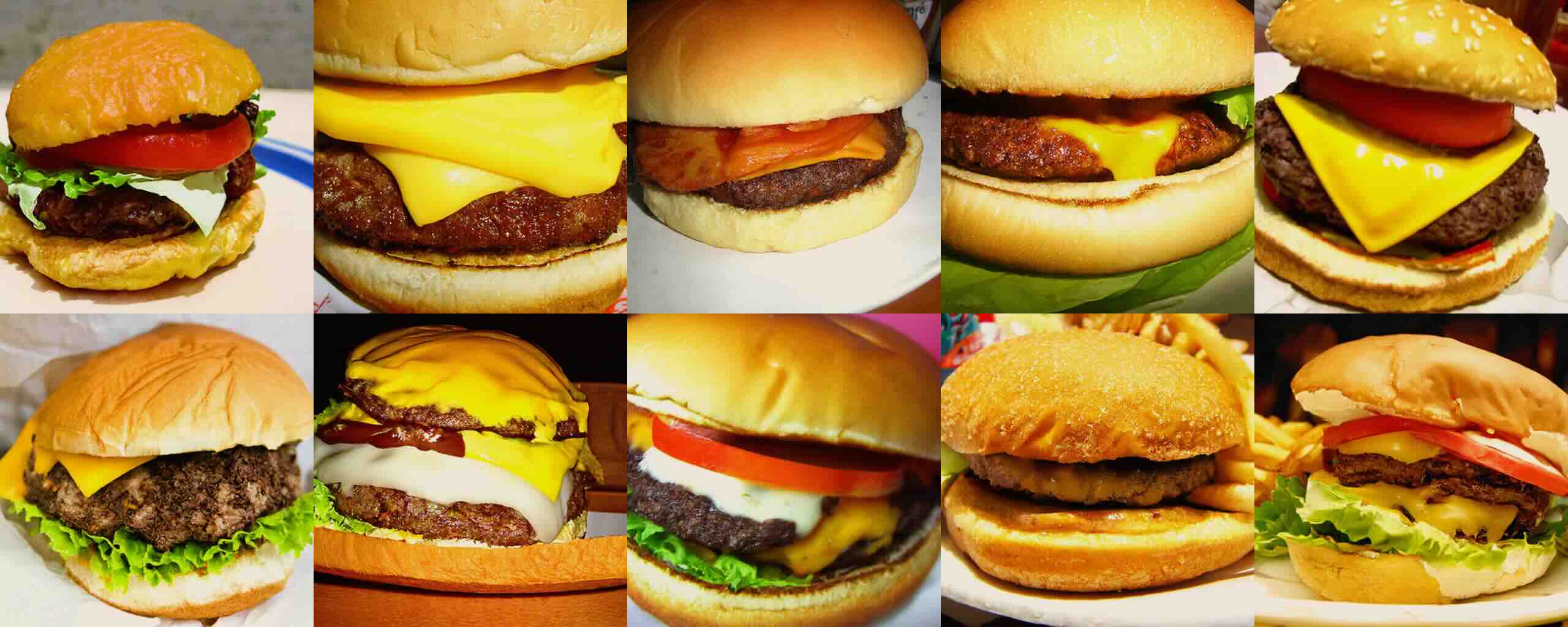}
        \end{minipage}
    \end{minipage}
    \caption{Uncurated samples. Class-to-image generation. Class: 1, 22, 89, 250, 356, 367, 429, 620, 812, 933. Setting: Heun 50 steps, CFG 4.0.}
    \label{app_fig:c2i_qualitative}
\end{figure}

\begin{figure}
    \centering
    \begin{minipage}[t]{\linewidth}
        \includegraphics[width=1.0\linewidth]{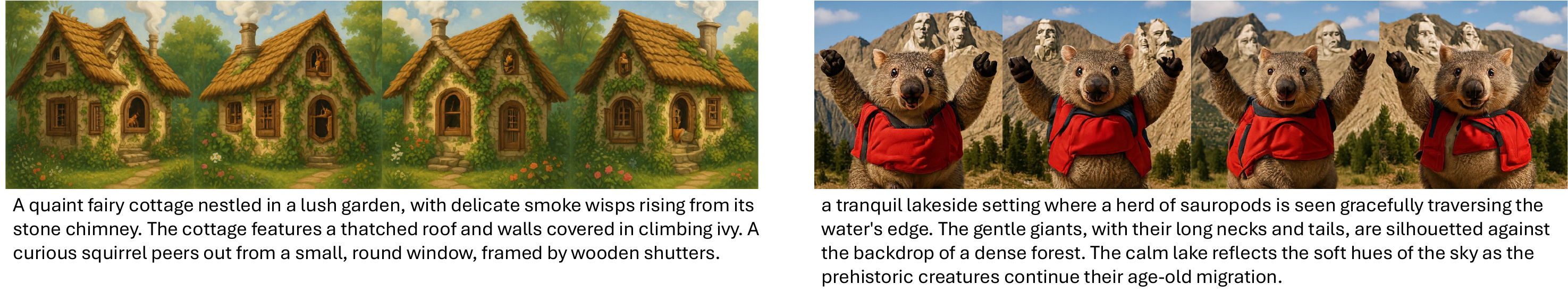}
    \end{minipage}
    \begin{minipage}[t]{\linewidth}
        \includegraphics[width=1.0\linewidth]{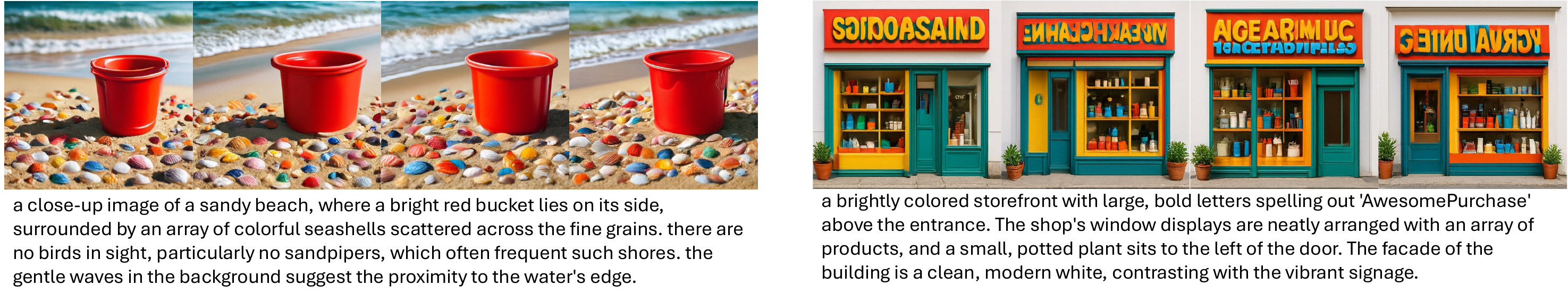}
    \end{minipage}
    \begin{minipage}[t]{\linewidth}
        \includegraphics[width=1.0\linewidth]{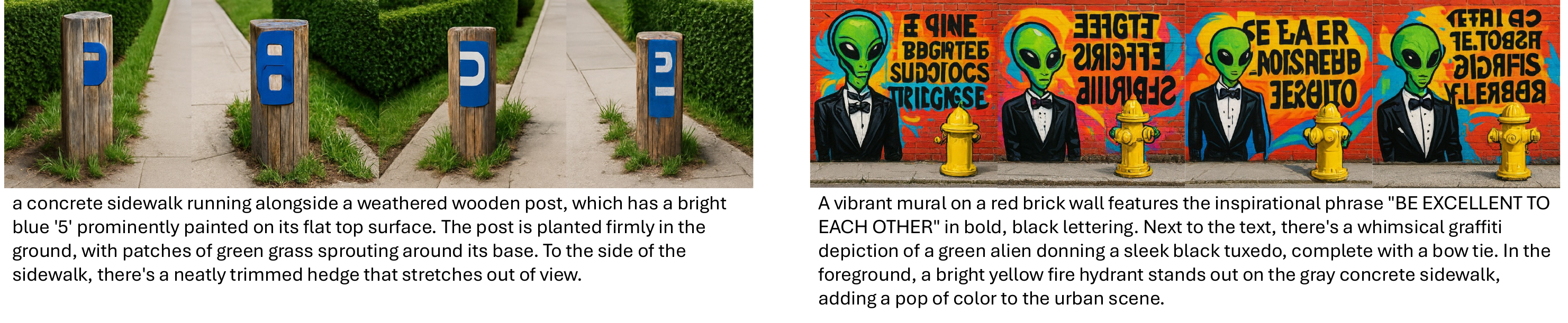}
    \end{minipage}
    \begin{minipage}[t]{\linewidth}
        \includegraphics[width=1.0\linewidth]{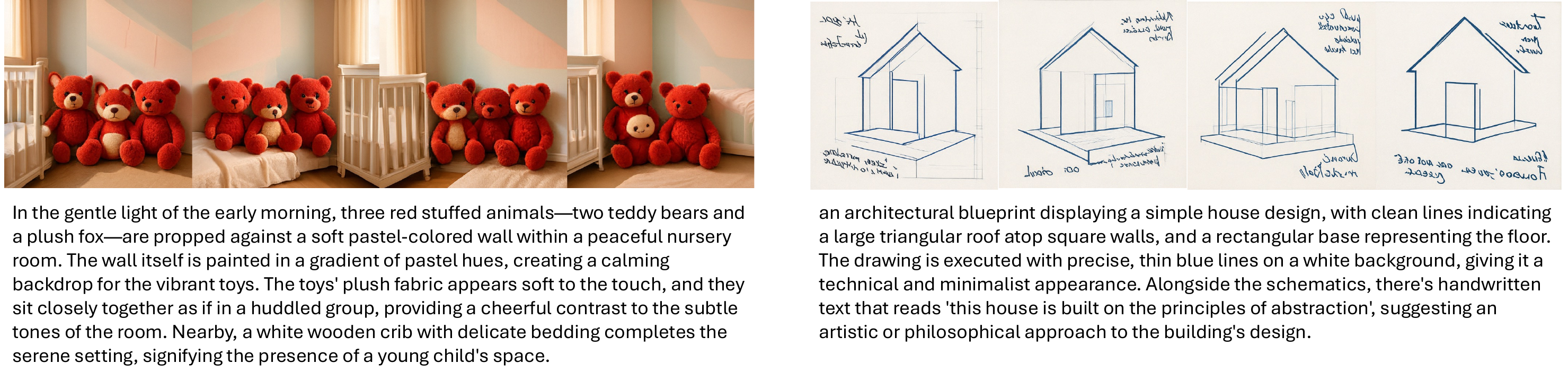}
    \end{minipage}
    \begin{minipage}[t]{\linewidth}
        \includegraphics[width=1.0\linewidth]{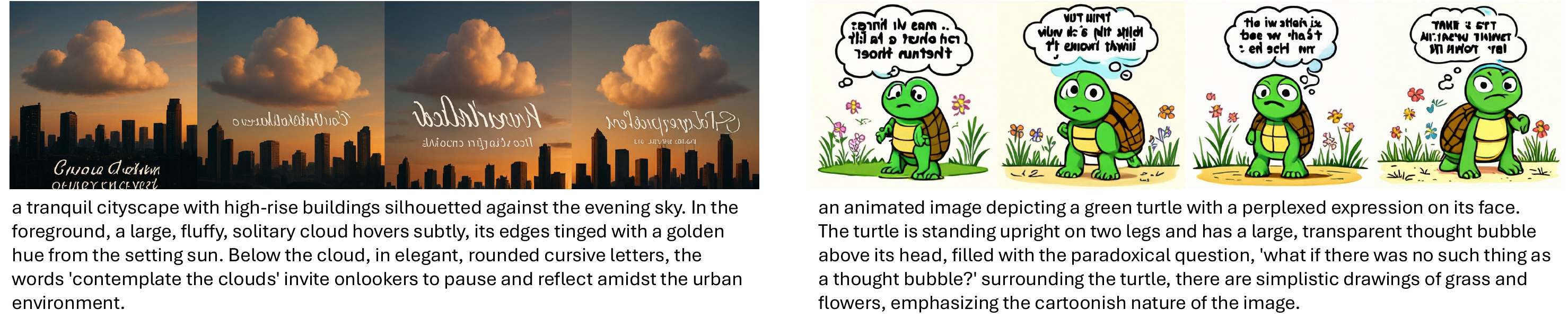}
    \end{minipage}
    \caption{Uncurated samples. Text-to-image generation.}
    \label{app_fig:t2i_qualitative}
\end{figure}

We display more qualitative results of our PiT-G/32 model trained on ImageNet-512 in Figure~\ref{app_fig:c2i_qualitative} and qualitative results of the text-to-image generation model PiT-L/32 in ~\ref{app_fig:t2i_qualitative}. We compress each image grid to around 100KB to avoid excessive supplementary file size.

